\documentclass[10pt]{article} 
\usepackage[accepted]{tmlr}

\usepackage{amsmath,amsfonts,bm}

\def\eqref#1{equation~\ref{#1}}

\def\1{\bm{1}}

\DeclareMathAlphabet{\mathsfit}{\encodingdefault}{\sfdefault}{m}{sl}
\SetMathAlphabet{\mathsfit}{bold}{\encodingdefault}{\sfdefault}{bx}{n}

\usepackage[utf8]{inputenc} 
\usepackage[T1]{fontenc}    
\usepackage{hyperref}       
\usepackage{url}            
\usepackage{booktabs}       
\usepackage{amsfonts}       
\usepackage{nicefrac}       
\usepackage{microtype}      
\usepackage{xcolor}         
\usepackage[table]{xcolor}
\usepackage{subcaption}
\usepackage{multirow}

\definecolor{commentcolor}{RGB}{110,154,155}   
\usepackage{graphicx}
\newcommand{\myparagraph}[1]{
\textbf{#1} ---
}

\usepackage{amsmath}
\usepackage[ruled,vlined]{algorithm2e}

\makeatletter
\def\@startauthor{\noindent\centering\normalsize}
\def\@endauthor{\par}
\def\@maketitle{\vbox{\hsize\textwidth
{\LARGE\bf\sffamily \@title\par}\vskip \aftertitskip
\if@accepted
    \if@preprint
        \@startauthor \@author \@endauthor
    \else
        \@startauthor \@author \\[0.8em] {\bf Reviewed on OpenReview:} \openreview \@endauthor
    \fi
\else
    \@startauthor Anonymous authors\\Paper under double-blind review \@endauthor
\fi
\vskip 0.3in minus 0.1in}}
\makeatother

\title{CDG-MAE: \underline{C}ross-view Masked Modeling using \underline{D}iffusion \underline{G}enerated Views}

\author{
  \textbf{Varun Belagali}\textsuperscript{1} \quad
  \textbf{Pierre Marza}\textsuperscript{5,6} \quad
  \textbf{Srikar Yellapragada}\textsuperscript{1} \quad 
    \textbf{Zilinghan Li}\textsuperscript{2} \quad
    \textbf{Tarak Nath Nandi}\textsuperscript{2,3} \quad
    \textbf{Ravi K Madduri}\textsuperscript{2,3} \quad
    \textbf{Joel Saltz}\textsuperscript{1} \quad
    \textbf{Stergios Christodoulidis}\textsuperscript{5,6} \quad \\
    \textbf{Maria Vakalopoulou}\textsuperscript{4,5,6} \quad
    \textbf{Dimitris Samaras}\textsuperscript{1} \quad
    \\[0.8em]
  {\small 
    \textsuperscript{1}Stony Brook University \quad
    \textsuperscript{2}Argonne National Laboratory \quad
    \textsuperscript{3}University of Chicago \quad
    \textsuperscript{4}Archimedes/Athena RC \quad
    \textsuperscript{5}Université Paris-Saclay, CentraleSupélec, Gustave Roussy, INSERM, IHU PRISM, Cancer Data Science Unit, France \quad \textsuperscript{6}Université Paris-Saclay, CentraleSupélec, MICS Laboratory, France
  }\\[0.5em]
  {\small
  \texttt{vbelagali@cs.stonybrook.edu} \\
  \url{https://github.com/cvlab-stonybrook/CDG-MAE}
  }
}

\def\month{08}  
\def\year{2026} 
\def\openreview{\url{https://openreview.net/forum?id=7XIymKIA0v}} 

\begin{document}

\maketitle

\begin{abstract}
Cross-view masked autoencoding has emerged as a powerful pretext task for learning dense correspondences, which are essential for applications such as video label propagation. The cross-view pretext task is modeled with a masked autoencoder, where a masked target view is reconstructed from an anchor view. However, acquiring effective training data remains a challenge - collecting diverse video datasets is costly, while simple image crops lack the necessary pose variations, underperforming video-based methods. This paper introduces CDG-MAE, a novel MAE-based self-supervised method that uses diverse synthetic views generated from static images via an image-conditioned diffusion model. We present a quantitative method to evaluate the local and global consistency of the generated views to choose the right diffusion model for cross-view self-supervised pretraining. These generated views exhibit substantial changes in pose and perspective, providing a rich training signal that overcomes the limitations of video and crop-based anchors. Furthermore, we enhance the standard single-anchor MAE setting to a multi-anchor masking strategy to increase the difficulty of the pretext task. CDG-MAE substantially narrows the gap to video-based MAE methods, while maintaining the data advantages of image-only MAEs.

\end{abstract}

\section{Introduction}

Masked Autoencoders (MAEs) learn rich visual representations by reconstructing randomly masked parts of an image from the remaining visible context \citep{he2022masked}. The paradigm of learning by reconstruction naturally extends to multi-view scenarios through cross-view correspondence learning ~\citep{weinzaepfel2022croco, weinzaepfel2023croco, gupta2023siamese, eymael2024efficient}.  These methods exploit the redundancy in captured information and the inherent 3D consistency across viewpoints as strong cues for learning to model dynamics, physics and semantics. A specific adaptation, cross-view masked auto-encoding, tasks a model to reconstruct a masked view of a scene from another anchor view. By learning to complete missing parts of scene representations, cross-view masked auto-encoding leads to strong vision models capable of understanding underlying scene semantics.

Training vision models to learn correspondences requires capturing multiple images of the scene, which in the real world can be costly. A common shortcut for {\em static environments} uses simulators to render diverse views of a scene ~\citep{weinzaepfel2022croco, weinzaepfel2023croco}. To model motion and perspective changes, the data itself must exhibit dynamic changes. Collecting videos is a good alternative ~\citep{gupta2023siamese}, but this comes with an acquisition cost as well as the more limited diversity of the scenes one can capture. For example, a video captures only a single motion scenario in a scene.

Given the large availability of 2D images, can we generate dynamic variations from images equivalent to those found in videos for correspondence learning? A simple approach is to emulate changes with augmentations such as image crops ~\citep{eymael2024efficient}, but the cropped view diversity is limited. As acknowledged by ~\citet{eymael2024efficient}, crops cannot introduce variations in pose, limiting the richness of learned inductive biases. 
Consequently, there is a need to develop methods that can use static images to derive richer, pose-variant transformations found in real-world dynamic scenes. While large-scale video datasets exist, collecting them is often resource-intensive or entirely unfeasible in specialized domains like medical imaging. Our goal is not to replace video data, but to maximize the utility of abundant, image-only datasets for effective cross-view pretraining.

\begin{figure*}[t]
    \centering
    \includegraphics[trim={0 2.5cm 0 0},clip,width=\linewidth]{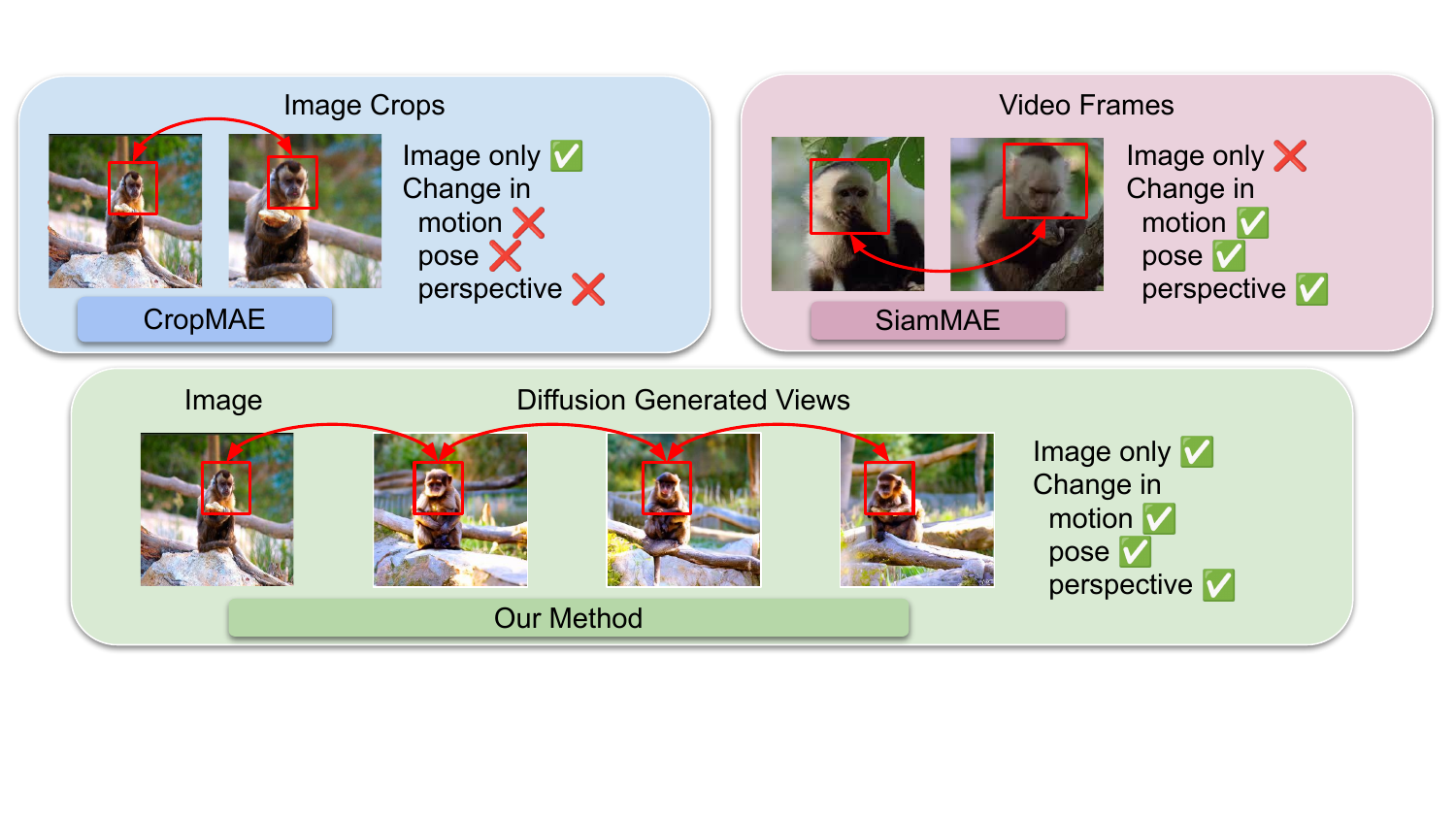}
    \vspace{-8mm}
    \caption{\textbf{CDG-MAE}: We train a vision encoder using cross-view masked autoencoding  between real and synthetic views generated by a diffusion model (~\citet{gen-sis}). These synthetic views preserve important scene information while introducing diverse dynamics.}
    \label{fig:Teaser}
\end{figure*}

Diffusion models perform well in image generation ~\citep{dhariwal2021diffusion, flux2024}, employing various conditioning mechanisms to guide the generation process ~\citep{rombach2022high, zhang2023adding}. Conditioning diffusion models with image embeddings enables the generation of diverse variations of an input image in a self-supervised way (S-LDMs ~\citet{rcg, gen-sis, ma2025learning}). Crucially, we observe that these variations can introduce different perspectives or diverse motions, equivalent to individual video frames (see Figure \ref{fig:Teaser}). However, in order to learn correspondences, the generated views should introduce local changes while maintaining global consistency. Currently, there are no quantitative tools to evaluate such properties.

We introduce \textit{CDG-MAE}, the first cross-view self-supervised learning method for correspondence learning, to train Masked Autoencoders using views generated from a diffusion model. 
However, the success of this strategy depends on the careful selection of the diffusion model as not all diffusion-generated views are useful for correspondence learning. In order for such views to provide a training signal analogous to video data, they must exhibit meaningful changes in pose and image location. To ensure this, we introduce quantitative consistency metrics to guide us in choosing the right diffusion model for generating these views. We observe a strong effect of these metrics on final performance. Traditional cross-view MAEs such as SiamMAE ~\citep{gupta2023siamese} and CropMAE ~\citep{eymael2024efficient} (single-anchor MAEs) typically encode a single unmasked anchor view. We extend the current single-anchor-view setting in cross-view self-supervised learning by using additional anchor views and applying an anchor-specific masking strategy (masking the anchor views) to increase the difficulty of the pretext task. 
Our contributions are as follows:

\textbf{(i) Diffusion-based view generation for MAE training.}  
We are the first to explore training cross-view MAEs using diffusion-generated views to address the limitations of video and image-crop based cross-view MAE methods.



\textbf{(ii) A method to evaluate the utility of diffusion generated-views for correspondence learning.} We develop quantitative metrics to evaluate local and global consistency between views. We demonstrate their effectiveness in choosing the right diffusion model for cross-view self-supervised learning.

\textbf{(iii) Multi-anchor masking as a novel MAE training paradigm.} We extend the standard single-anchor MAE setting to a multi-anchor framework. Having multiple anchors allows for anchor masking, which creates a more challenging and effective pretext task.

We show that CDG-MAE, trained with diffusion-generated data and our multi-anchor setting, achieves substantial improvements over state-of-the-art MAE methods reliant on image crops and narrows the performance gap with video-based approaches.

\section{Related work}

\myparagraph{Self-supervised learning} Masked Image Modeling (MIM) is a self-supervised learning paradigm that masks part of the input visual data and trains models to predict the masked parts using visible parts ~\citep{he2022masked, bao2021beit,xie2022simmim, wei2022masked, tong2022videomae, jepa}. Specifically, Masked Autoencoders (MAE ~\citet{he2022masked}) divide an image into patches, and mask some of them. An encoder extracts features from visible patches only. The encoder features and appended mask tokens with positional encoding are used to decode the patch pixel values. With a sufficiently high masking ratio, the encoder learns robust visual features for downstream discriminative tasks (classification, segmentation, object detection). VideoMAE ~\citep{tong2022videomae} pretrains on videos integrating multiple frames.  Another class of SSL methods are view-invariant methods which use two augmentations of the same image, and train the model to match the global/local features between augmentations ~\citep{chen2020simple, dinov1, grill2020bootstrap, he2020momentum, zhou2021ibot, oquab2023dinov2, bardes2022vicregl}. In this work, we specifically focus on MAE as a SSL framework due to its efficiency and modularity.


\myparagraph{Cross-view self-supervised learning} learns visual features that match cross-views either for video ~\citep{gupta2023siamese, eymael2024efficient} or 3D ~\citep{weinzaepfel2022croco, weinzaepfel2023croco} downstream tasks. These works use Siamese Masked Autoencoders to learn cross-view correspondences. Pretraining employs two views: anchor and target. The masked target image passes through the encoder, then the decoder reconstructs the masked patches. The anchor view passes through the encoder independently, without masking. To facilitate cross-view learning, the decoder reconstructs the target view by cross-attending to the encoder features of the anchor view. A high masking ratio forces the encoder to learn features that match patches of the anchor view to the target view. Siamese Masked Autoencoders (SiamMAE~\cite{gupta2023siamese}) extract target and anchor as two different video frames. Given object motion, view point change, and pose change in video, SiamMAE visual features are suitable for label propagation downstream tasks: video object propagation ~\citep{davis-vos}, semantic part propagation ~\citep{vip}, and pose propagation ~\citep{jhmdb}. Cropped Siamese Masked Autoencoder (CropMAE~\citet{eymael2024efficient}) extends SiamMAE, extracting the two views from two crops of the same image, obviating the need of video pretraining. CropMAE performs worse on tasks like pose propagation, as the anchor and target views have limited pose changes.

\myparagraph{Diffusion models} generate realistic images due to breakthroughs in conditioning \citep{ho2022classifier,zhang2023adding}, architecture \citep{peebles2023scalable, esser2024scalingrectifiedflowtransformers} and sampling \citep{song2020denoising, lu2023dpmsolverfastsolverguided}. Latent Diffusion Models (LDMs~\citet{rombach2022high}) efficiently train a diffusion model in a compact VAE latent space instead of pixel space. While diffusion models are commonly conditioned on explicit signals such as class labels and text captions, recent works ~\citep{gen-sis, rcg, ma2025learning, Graikos_2024_CVPR} train conditional diffusion models in a self-supervised  way (S-LDMs). These approaches first train an image encoder using view-invariant self-supervised learning methods ~\citep{dinov1, moco} to learn image embeddings. The diffusion model is then conditioned on the output of the frozen encoder. 

Diffusion models are increasingly used for data augmentation, including self-supervised settings. \citep{tian2024learning, tian2024stablerep} use Stable Diffusion to generate augmentations, whereas, approaches like Gen-SIS \citep{gen-sis} train the diffusion model on the same dataset used for the main SSL task. We experiment with three different self-supervised image-conditioned diffusion models: Gen-SIS~\citep{gen-sis}, RCG~\citep{rcg}, and Lumos~\citep{ma2025learning}. Gen-SIS and RCG are pretrained on ImageNet-1K, while Lumos is pretrained on 190M open-source images.

\begin{figure}
    \centering
    \includegraphics[width=\linewidth]{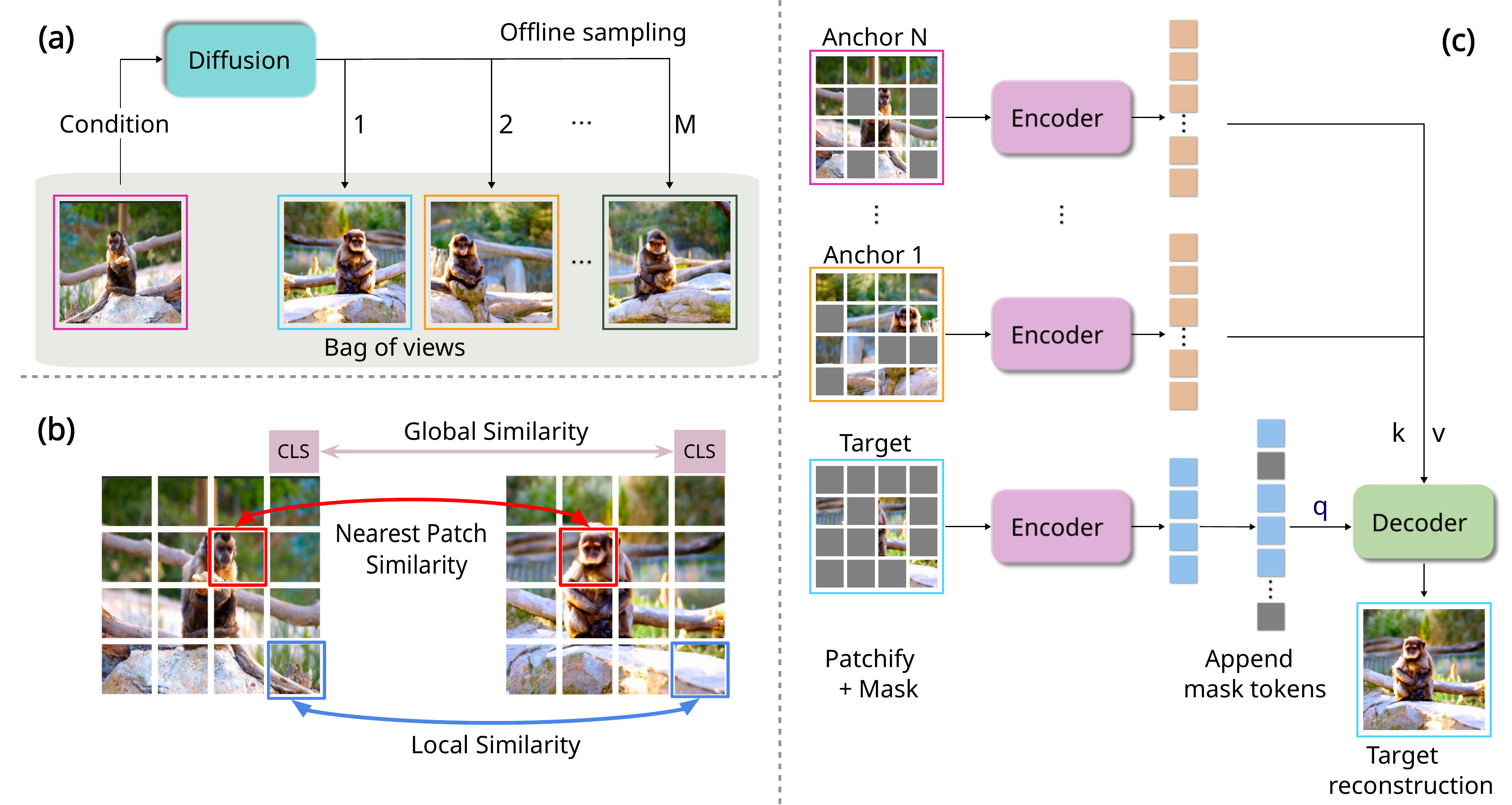}
    \vspace{-8mm}
    \caption{\textbf{Overview of CDG-MAE}: \textbf{(a)} For every real image, we generate $M$ views using an off-the-shelf S-LDM \citep{gen-sis}. \textbf{(b)} We develop quantitative metrics to evaluate local and global consistencies between view pairs. \textbf{(c)} We develop a multi-anchor framework to train cross-view MAE. Having multiple anchors allows for anchor masking, which creates a more challenging pretext task.} 
    \label{fig:method}
\end{figure}

\section{Method}

Our approach, CDG-MAE, consists of three stages: 1) Bag of views generation. 2) Quantitative view evaluation and 3) Cross-view MAE training. Figure~\ref{fig:method} describes the overall pipeline of CDG-MAE.

\subsection{Bag of views} 
Given an image-only dataset (e.g., ImageNet), we generate \textit{M} alternative views of the scene depicted by each real image via an 
off-the-shelf image-conditioned diffusion model. We mainly use a Self-supervised Latent Diffusion Model (S-LDM), pretrained with ImageNet self-supervision ~\citep{gen-sis}. S-LDM follows the Latent Diffusion Model~\citep{rombach2022high} architecture and is conditioned with an image encoder pretrained with view invariant SSL ~\citep{dinov1} and frozen during LDM training. We observe that S-LDM generates diverse views of an input image with motion, pose, and perspective variations, mimicking changes between video frames (Figure \ref{fig:Teaser}). These offline generated views, along with the original image, are the \textit{bag of views} for that input, avoiding online sampling during MAE training (with a one-time cost of 135 ms per real image). 

\subsection{Evaluating consistency between view pairs}
\label{sec:metrics}

Intuitively, an ideal pair of views $(V_1, V_2)$ for correspondence learning should feature the same set of objects undergoing transformations in motion, pose, and perspective. Such views must therefore exhibit local variations that reflect these transformations while maintaining  global consistency. To measure these properties, we developed consistency metrics. Figure~\ref{fig:method} (b) provides a visual illustration.

Let $f_*(\cdot)$ and $f_i(\cdot)$ denote functions that extract a transformer-based global and local embedding at a spatial location $i$, respectively for a single input $V_1$. In our case, $f_*(V_1)$ is the $[CLS]$ token and $f_i(V_1)$ is the $i^{th}$ patch token extracted from a ViT encoder. We use the pretrained ViT-B/16 MAE ~\citep{he2022masked} to extract both patch tokens and CLS tokens. Let us assume there are $L$ distinct spatial locations, indexed from $1$ to $L$. Using this information, we calculate the:

\myparagraph{Global Similarity (GS)} Measures the overall semantic coherence between views $V_1$ and $V_2$. It is defined as the cosine similarity ($sim(\mathbf{u}, \mathbf{v})$) between their respective global embeddings, $f_*(V_1)$ and $f_*(V_2)$. 
High GS is desirable, indicating that global semantic content is preserved.

\myparagraph{Local Similarity (LS)} It is the average cosine similarity between local embeddings $f_i(V_1)$ and $f_i(V_2)$ from $V_1$ and $V_2$ respectively, at each identical spatial location $i$. A low LS indicates change in motion, pose, and perspective between views.

\myparagraph{Nearest Patch Similarity (NPS)} Provides a measure of global consistency, especially in the presence of transformations. For each local  embedding $f_i(V_1)$ in $V_1$, we identify its nearest neighbor (most similar) among all local  embeddings $\{f_j(V_2)\}_{j=1}^L$ from $V_2$. NPS is then calculated as the average of these maximum similarity scores across all $L$  locations in $V_1$. Even with significant changes in motion, pose, or perspective, a high NPS is expected if the views remain globally coherent.

\begin{align}
    GS(V_1, V_2) &= sim(f_*(V_1), f_*(V_2)) \\
    LS(V_1, V_2) &= \frac{1}{L} \sum_{i=1}^{L} sim(f_i(V_1), f_i(V_2)) \\
    NPS(V_1, V_2) &= \frac{1}{L} \sum_{i=1}^{L} \left( \max_{j \in \{1, \dots, L\}} sim(f_i(V_1), f_j(V_2) \right)
\end{align}

Table~\ref{tab:metrics} presents the evaluation of quantitative metrics on several types of view pairings: (i) video frames, (ii) synthetic views generated by the S-LDM, (iii) k-nearest neighbor (k-nn) image pairs from the training data (derived from cosine similarity in S-LDM conditioning encoder space), and (iv) random image pairs. We select video frames following the SiamMAE data loading protocol: a primary frame is chosen at random, and a second frame is randomly sampled with a delta of 4 to 48 frames. The result demonstrates that S-LDM generated views are much closer to video frames than k-nn images or random pairs. This indicates that S-LDM, conditioned on static real images, can mimic characteristics of video data, making it ideal for correspondence learning. In Section~\ref{sec:choice_of_diffusion_model}, we demonstrate the effectiveness of the above metrics in choosing the right diffusion model for view generation.

\begin{table}[!htbp]
\centering
{
\caption{Quantitative evaluation of diffusion generated views when compared to video frames, k-nn images, and random pair of images. We sample random 5000 images from ImageNet~\citep{deng2009imagenet} and 5000 pair of video frames from Kinetics-400~\citep{kay2017kinetics} for calculation.}
\label{tab:metrics}
\resizebox{0.7\textwidth}{!}{%
\begin{tabular}{c|ccc}
\toprule
   Views & Global Sim. ($\uparrow$) & Local Sim. ($\downarrow$)  & Nearest Patch Sim.($\uparrow$ )  \\

   \midrule
    Video frames &  0.992	 & 0.389  & 0.884   \\
    \rowcolor{blue!10}
    \hline
    Diffusion (S-LDM) &  0.992 & 0.377   & 0.795  \\
    K-nn images &  0.951	 & 0.301	 & 0.719   \\
    Random images &  0.892	 & 0.175	 & 0.600   \\

\end{tabular}}}
\end{table}

\subsection{CDG-MAE overall design and training strategy}
In this section, we explain the overall design and training methodology for CDG-MAE: cross-view masked autoencoders using diffusion generated views. Consistent with existing works on cross-view MAE ~\citep{gupta2023siamese,eymael2024efficient}, the pretext task is the reconstruction of randomly masked patches in a target,  using visible target patches and anchor views. The architecture is an encoder-decoder Vision Transformer (ViT), where the target and each anchor are independently processed by a weight-shared ViT encoder. Subsequently, the decoder appends mask tokens to the visible target tokens, and reconstructs the content for these masked patches. This reconstruction is conditioned on visible target tokens through self-attention and on anchor tokens via cross-attention.

In the remainder of this section, we denote the encoder and decoder as $e_\theta$ and $d_\psi$ respectively. We define image patchification operator as $\rho(\cdot)$, masking operator as $m(\cdot, \text{ratio})$, and concatenation as $[\cdot; \cdot]$. The masking ratios for the target and anchor views are denoted by $r_t$ and $r_a$, respectively.

\myparagraph{Encoding Target} The target image $T \in \mathbb{R}^{H \times W \times 3}$  is first patchified into a sequence of $N_t = (H/P) \times (W/P)$ non-overlapping patches, each of size $P \times P \times 3$. We then flatten these patches into a 1D sequence, and apply random masking using a high target masking ratio ($r_t$). We discard the masked patches $\tilde{T}_v$, and process the visible patches $T_v$ through the encoder $e_\theta$ to obtain encoder target representations $T'_v$.

\begin{gather}
    T_p = \rho(T), \quad T_v, \tilde{T}_v = m(T_p, r_t)  \label{eq:patchmasktarget} \\
    T'_v = e_\theta(m(T_p, r_t); \theta)  \label{eq:targetencoder}
\end{gather}

\myparagraph{Multi-anchor and anchor-masking} Traditional cross-view MAEs such as SiamMAE ~\citep{gupta2023siamese} and CropMAE ~\citep{eymael2024efficient} typically encode a single unmasked anchor view. In CDG-MAE, we propose leveraging multiple anchor views, $\{A^k\}^{N}_{k=1}$, sampled from the "bag of views" (where $N$ is the number of anchors). Furthermore, we introduce \textit{anchor masking}: each anchor view $A^k$ is independently masked with a specific anchor masking ratio $(r_a)$, allowing for fine-grained control over the difficulty of the pretext task. Higher $N$ can provide the decoder with richer contextual information, simplifying target reconstruction. Conversely, applying anchor masking makes the task more challenging by reducing the visible information from each anchor. We demonstrate that an optimal balance between $N$ and $r_a$ enhances representation learning.

Similar to target encoding, each anchor $A^k$ is patchified and masked. We discard masked anchor patches, and pass the visible patches through the weight-shared encoder $e_\theta$ (Siamese-style encoding) to obtain anchor tokens $A'^{k}_v$. We then concatenate the output tokens from all $N$ anchors to form a aggregated anchor representation $A'_v$

\begin{align}
    A'^{k}_v &= e_\theta(m(\rho(A^k),r_a); \theta),  &\forall k \in \{ 1, \ldots, N\} \\
    A'_v &= [A'^{1}_v; A'^{2}_v; ..; A'^{N}_v] 
    \label{eq:anchorencoder}
\end{align}

\myparagraph{Target reconstruction} The input to the decoder $d_\psi$ is the sequence $T_a$, which is a concatenation of encoder target representations $T'_v$ and mask tokens $M_{\tilde{T}_v}$. The decoder self-attends to all tokens within $T_a$, and cross-attends to the aggregated anchor representation $A'_v$, allowing it to leverage information across anchor views to predict the masked target patches.

Our multi-anchor setting encourages the encoder $e_\theta$ to learn features that are robust for matching across a diverse set of views - beyond just two views as in prior work ~\citep{ gupta2023siamese, eymael2024efficient}.  We apply a reconstruction loss (MSE) between masked target patches $\tilde{T}_v$ and decoder predictions $T_r$, following prior work. 

\begin{align}
    T_a &= [T'_v, M_{\tilde{T}_v}]   \\
    T_r &= d_\psi(T_a, A'_v; \psi)   \\
    \mathcal{L}(T_r, T_p) &= \frac{1}{|\tilde{T}_v|}\left\lVert T_r - \tilde{T}_v \right\rVert^2_2
    \label{eq:targetrecon}
\end{align}

\section{Experimental setting}
\label{sec:experimental_setting}
\myparagraph{Bag of Views creation} For each real image in the ImageNet-1K ~\citep{deng2009imagenet} training dataset, we generate $M=4$ random synthetic views using the pretrained checkpoint of S-LDM ~\citep{gen-sis}. The real image along with generated views are treated as the \textit{bag of views}. The generation is done in offline mode and stored on the disk before training CDG-MAE. Following ~\citep{gen-sis}, we use a classifier-free guidance weight~\citep{ho2022classifier} of 6  and 50 DDIM ~\citep{song2020denoising} steps for sampling.

\myparagraph{Training} We utilize the official codebase of CropMAE ~\citep{eymael2024efficient} and closely follow their setting. By default, we use a ViT-S/16 encoder and a four-layer decoder. Each decoder block has an embedding dimension of 256, and contains cross-attention, feedfoward and self-attention modules. We train for 100 epochs on ImageNet-1K with a base learning rate of $1.5 \times 10^{-4}$ and batch size of 2048. 

From the \textit{bag of views} (containing $M$ views), one image is randomly chosen as the target and $N$ additional images as anchors ($N<M$). 
We use a target masking ratio $r_a=90\%$ . In the multi-anchor setting, we apply uniform anchor masking ratio across all anchors, with each anchor masked independently and randomly. We also investigate the impact of training with a reduced patch size by training both CropMAE and CDG-MAE with a ViT-S/8 backbone for 100 epochs. More training details are provided in Appendix ~\ref{sec:sup_training_details}.

\myparagraph{Downstream evaluation} Following existing works ~\citep{gupta2023siamese, eymael2024efficient} we evaluate  cross-view pretraining using three label propagation tasks in videos - 1) DAVIS-2017 video object segmentation ~\citep{davis-vos} , 2) VIP  semantic part propagation ~\citep{vip}, and 3) JHMDB human pose propagation ~\citep{jhmdb}. In label propagation, we are provided with the annotation of the first frame and the task is to propagate the label to all frames by computing the similarity (correspondence) between patches of frames. The evaluation is done in a training-free manner using the pretrained encoder following the setting of ~\citep{gupta2023siamese, eymael2024efficient}. For all downstream tasks, higher values indicate better performance. More details are provided in the Appendix ~\ref{sec:sup_down_eval}.

\section{Results}
\label{sec:results}
We first discuss the design choices of CDG-MAE in Sec \ref{sec:res_target},\ref{sec:choice_of_diffusion_model}, \ref{sec:res_anchors} and then compare with MAE-based methods in Sec \ref{sec:res_mae16}. We present results on training with a smaller patch size in Sec \ref{sec:res_scaling}. 

\subsection{Target selection and masking}
\label{sec:res_target}

To investigate the influence of target selection, we employ a simple single-anchor configuration where the anchor view is unmasked $(r_a = 0)$ and with a high target masking ratio $(r_t = 90\%)$. We evaluate three strategies for selecting the target view: 1) always using the real image, 2) always using a diffusion-generated view or 3) randomly choosing between real and generated. We observe in Table \ref{tab:res_target} that always using the real image as target and the random choice strategy yield comparable, strong performance.  Hence, we adopt the random choice selection as the default strategy for CDG-MAE. 

In Table \ref{tab:res_target}, we also tested using k-nearest neighbor ($k=5$) image pairs for anchor and target, and observe that it underperforms our default strategy. While k-nn images might share global feature similarity, Table \ref{tab:metrics} indicates they lack the high Nearest Patch Similarity found in video frames.

\begin{table}
\centering
  \caption{Effect of target view selection (real, generated, or $k$-nn images). It is optimal to choose real or generated image as the target view.}
  \label{tab:res_target}
  {
\begin{tabular}{c|ccc}
\toprule
   Target View & DAVIS & VIP & JHMDB \\
     & $\mathcal{J} \& \mathcal{F}_m$  & mIoU & PCK\@0.1 \\
   \midrule
   
    Always Real &  61.3	 & 37.1	 & 46.8 \\
    Always Generated &  60.0 & 37.1 & 46.5	 \\
    \rowcolor{blue!10}
    Real or Generated &  61.2	& 37.6 & 46.5 \\
    k-nn image & 60.5	 & 36.0	 & 46.5 \\

\end{tabular}}
\end{table}

Next, in Table \ref{tab:res_targetmasking}, we investigate the effect of target masking ratio ($r_t$).  Unlike vanilla MAE (which uses $r_t = 75\%$), cross-view MAEs such as ~\citep{eymael2024efficient, gupta2023siamese} typically employ higher ratios ($\ge$90\%). Lower ratios (e.g., 75\%) can encourage the target to reconstruct itself, thereby hindering correspondence learning from anchor views. CropMAE uses a very high ratio ($r_t =98.5\%$) to keep the task challenging under high information redundancy between anchor and target cropped from the same image. In CDG-MAE, such high ratio performs poorly. This can be attributed to the greater visual variations between diffusion generated anchor views and the target image compared to simple image crops. In our case, target reconstruction requires more context. We observe that a balanced masking ratio of $r_t=90\%$ yields optimal performance.

\begin{table}
\centering
  \centering
  \caption{Effect of target masking ratio ($r_t$). A balanced target masking ratio of $r_t = 90\%$ yields the best performance.}
  \label{tab:res_targetmasking}
  {
\begin{tabular}{c|ccc}
\toprule
    Masking  & DAVIS & VIP & JHMDB \\
     Ratio (\%) & $\mathcal{J} \& \mathcal{F}_m$ & mIoU & PCK\@0.1 \\
   \midrule
   
    75 &  60.0	 & 34.9	 & 46.2 \\
    \rowcolor{blue!10}
    90 &  61.2 & 37.6 & 46.5 \\
    98.5 &  60.7 & 35.5 & 44.4 \\
 
\end{tabular}}
\end{table}

\subsection{Diffusion model selection}
\label{sec:choice_of_diffusion_model}
In this section, we experiment training CDG-MAE (single-anchor setting) using views generated from three diffusion models pretrained in a self-supervised manner (S-LDM): Gen-SIS~\citep{gen-sis}, RCG~\citep{rcg}, and Lumos~\citep{ma2025learning}. Gen-SIS and RCG are trained on ImageNet-1K, while Lumos is trained on a large collection of 190M images from open source datasets. Table~\ref{tab:choice_of_diffusion_model} shows the performance of CDG-MAE when trained with views from different diffusion models, along with our proposed consistency metrics (GS, LS, NPS). We also report the metrics on Kinetics video frames for reference. The table shows that views generated from Gen-SIS and Lumos are considerably closer to video frames than RCG-generated views in terms of GS, LS, and NPS. Similarly, we observe that training with views from either Gen-SIS or Lumos offers higher performance than training with RCG. This finding demonstrates the importance of our proposed metrics in identifying the right diffusion model for cross-view pretraining. Based on these metrics, we chose Gen-SIS over RCG as our default diffusion model. Since Lumos is trained on a larger scale dataset than ImageNet-1K, we do not use it in our experiments to avoid potential data leakage effects. We did not include image-to-video diffusion models in our study because it would require immense computational resources (tens of thousands of A100 hours) to generate videos at ImageNet scale (see Appendix \ref{sec:sup_video_diffusion}).

\begin{table}[!htbp]
\centering
{
\caption{CDG-MAE performance with different diffusion models (S-LDMs) and corresponding consistency metrics (GS, LS, NPS). Video frame metrics provided as reference. Our proposed consistency metrics strongly effect the performance.}
\label{tab:choice_of_diffusion_model}
\resizebox{\textwidth}{!}{%
\begin{tabular}{c|ccc|ccc}
\toprule
    Diffusion Model & DAVIS  & VIP & JHMDB & Global Sim. ($\uparrow$) & Local Sim. ($\downarrow$)  & Nearest Patch Sim.($\uparrow$ )  \\
    \midrule
    \rowcolor{blue!10}
    Gen-SIS (~\cite{gen-sis}) & 61.2 & 37.6 & 46.5 & 0.992 & 0.377 &  0.795\\
    RCG (~\cite{rcg}) & 57.4 & 34.8 & 43.7 & 0.955 & 0.308 & 0.738 \\
    Lumos (~\cite{ma2025learning}) & 61.9 & 37.7 & 47.3 & 0.995 & 0.376 & 0.812 \\
    \midrule
    Video frames & NA & NA & NA & 0.992 &  0.389 & 0.884 \\
    

\end{tabular}}}
\end{table}

\subsection{Multi-anchor and anchor-masking} 
\label{sec:res_anchors}

In this section, we explore the extension from a single-anchor to a multi-anchor framework. The latter enables us to introduce anchor masking, a technique not explored in prior methods such as SiamMAE ~\citep{gupta2023siamese} and CropMAE ~\citep{eymael2024efficient}.

Table~\ref{tab:res_multi-anchor} demonstrates that transitioning from a single anchor to two anchors substantially improves performance. With two anchors, the decoder can learn more robust correspondences by matching target patches across both views. Within the two-anchor configuration, applying an anchor masking ratio of 25\% or 50\% further enhances the results. Ultimately, we achieve optimal performance using three anchors with a 25\% anchor masking ratio, as presented in Table~\ref{tab:res_multi-anchor}. This highlights the need for anchor masking when increasing the number of anchors to avoid providing too much information to the decoder, which would greatly reduce the task difficulty.

Next, we discuss the associated training computational complexity for multi-anchor and anchor-masking
(evaluation-time inference complexity is the same for all models). GLOPs are calculated using a full forward pass up to the loss calculation with a single input training sample. Adding more anchors without anchor masking increases the computational complexity as it results in more tokens being processed by the encoder and decoder. However, introducing anchor masking not only improves performance, but can also help in reducing GLOPs.  This is since the masked tokens are dropped at the input and hence do not add complexity, effectively decreasing the number of tokens processed by the encoder for each anchor, and finally the number of tokens used in cross-attention by the decoder. As presented in Table~\ref{tab:res_multi-anchor}, $N=2$ anchors and $r_a=50$\%  can maintain almost the same number of FLOPs as the single-anchor setting and outperforms it in downstream tasks. Further, to make the most out of multiple anchors, our best model uses $N=3$ anchors and $r_a=25$\%. Finally, for $N=4$ we noticed no further improvement and even a small decrease in performance. For this reason, we performed only one seed run for $N=4$.

\begin{table}[!htbp]
\centering
{
\caption{Effect of multiple anchors and anchor masking ($r_a$). Multi-anchor training improves performance, and anchor masking offers control over pretext task difficulty. We report mean $\pm$ std across 3 pretraining runs (seeds). We also report GLOPs to compare complexity.}
\label{tab:res_multi-anchor}
\resizebox{0.6\columnwidth}{!}{
\begin{tabular}{c|c|ccc|c}
\toprule
    Num. of  & Anchor  & DAVIS & VIP & JHMDB & GFLOPs \\
     Anchors ($N$) & Masking ratio ($r_a$) & $\mathcal{J} \& \mathcal{F}_m$ & mIoU & PCK\@0.1 \\
   \midrule
   \rowcolor{blue!10}
    1 & 0 & 61.2$ \pm$0.0 & 37.6 $\pm$0.4 & 46.5 $\pm$0.3  & 6.0 \\
   \midrule
    2 & 0 & 62.0 $\pm$0.1	 & 37.6 $\pm$0.1  & 47.1 $\pm$0.2 & 10.4 \\
    2 & 25\% &  62.4 $\pm$0.2	 & 38.0 $\pm$0.3 & 47.3 $\pm$0.1 &  8.3	 \\
    \rowcolor{blue!10}
    2 & 50\% &  62.1 $\pm$0.1	 & 38.1 $\pm$0.2 & 47.8 $\pm$0.1 & 6.1 \\
   \midrule

   \rowcolor{blue!10}
     3 & 25\% &  62.6 $\pm$0.1	 & 38.1 $\pm$0.1	 & 47.8 $\pm$0.2  & 11.6 \\
     3 & 50\% &  62.0 $\pm$0.4 & 37.4 $\pm$0.3 & 47.5 $\pm$0.2 & 8.3 \\
   \midrule
     4 & 25\% &  62.3	 & 37.6	 & 47.6  & 14.9 \\
     4 & 50\% &  61.7 & 37.4 & 47.6 & 10.6 \\
\end{tabular}}}
\end{table}

\subsection{Comparison with other Masked Autoencoders}

\label{sec:res_mae16}

We compare CDG-MAE with multiple MAE baselines. These include the vanilla MAE ~\citep{he2022masked}, video-based MAEs - Video-MAE ~\citep{tong2022videomae} and MAE-ST ~\citep{feichtenhofer2022masked}, and other cross-view MAEs such as CropMAE ~\citep{eymael2024efficient}, SiamMAE ~\citep{gupta2023siamese}, and CroCo ~\citep{weinzaepfel2022croco, weinzaepfel2023croco}. While we include CroCo as a baseline, it is not a direct comparison to our method as it uses 3D data during pretraining. We provide CroCo results primarily for broader benchmarking context. We evaluate two CDG-MAE variants: CDG-MAE-a1 (single unmasked anchor) and CDG-MAE-a3 (three anchors with 25\% masking).

\begin{table}[!htbp]
\centering
{
\caption{Comparison of CDG-MAE with other MAE based methods. a1 and a3 refer to single anchor and three anchors with 25\% anchor masking respectively. $\dagger$ refers to our reproduction on ImageNet-1K. The best and second-best results are highlighted in \textbf{Bold} and \underline{Underline}.}
\label{tab:res_vits16}
\resizebox{1\columnwidth}{!}{
\begin{tabular}{c|ccc|ccc|c|cc}
\toprule
   Method & Arch & Dataset & Epochs & \multicolumn{3}{c|} {DAVIS} & VIP & \multicolumn{2}{c}{JHMDB} \\
    & & & & $\mathcal{J}  \& \mathcal{F}_m$ ($\uparrow$) & $\mathcal{J}_m$ ($\uparrow$) & $\mathcal{F}_m$ ($\uparrow$) & mIoU ($\uparrow$) & PCK\@0.1 ($\uparrow$) & PCK\@0.2 ($\uparrow$) \\
    \midrule
    MAE (~\cite{he2022masked}) &  ViT-B/16 & ImagNet-1K & 1600 &   53.5 & 52.1 & 55.0 & 28.1 & 44.6 & 73.4  \\ 
    Video-MAE (~\cite{tong2022videomae}) &  ViT-S/16 & Kinetics-400 & 800 &   39.3 & 39.7 & 38.9 & 23.3 & 41.0 & 67.9  \\ 
    MAE-ST (~\cite{feichtenhofer2022masked}) &  ViT-L/16 & Kinetics-400 & 800 &   54.6 & 55.5 & 53.6 & 33.2 & 44.4 & 72.5   \\
    
    \midrule
    CroCov1 (~\cite{weinzaepfel2022croco})  &  ViT-B/16  & Habitat  &  400 & 55.9  & 52.9  & 58.9  & 31.3 & 42.3	  & 70.6  \\
    CroCov2 (~\cite{weinzaepfel2023croco})  &  ViT-B/16  & Habitat + Real  &  100 & 56.5 & 53.0  &  60.0  & 32.1 & 44.6  &  72.8 \\
    CroCov2 (~\cite{weinzaepfel2023croco}) &  ViT-L/16  & Habitat + Real  &  100 &  57.9 & 54.4  &  61.4 & 31.7 & 43.4	  & 71.3 \\
    SiamMAE (~\cite{gupta2023siamese}) & ViT-S/16 & Kinetics-400 & 2000 & \underline{62.0} & \textbf{60.3} & 63.7 & 37.3 & \underline{47.0} &  \underline{76.1} \\
    CropMAE $\dagger$ (~\cite{eymael2024efficient}) & ViT-S/16 & ImagNet-1K & 100 & 59.7 & 56.9 & 62.5 & 33.8 & 43.9 & 72.3 \\
    \rowcolor{blue!10}
    CDG-MAE-a1 &  ViT-S/16 & ImagNet-1K & 100 & 61.2 & 58.1 &	\underline{64.3} & \underline{37.6} & 46.5 & 75.5 \\
    \rowcolor{blue!10} 
    \rowcolor{blue!10}
    CDG-MAE-a3 &  ViT-S/16 & ImagNet-1K & 100 &   \textbf{62.6} & \underline{59.7} & \textbf{65.5} & \textbf{38.1} & \textbf{47.8}  &  \textbf{76.3}  \\
    
\end{tabular}}}
\end{table}

Results in Table \ref{tab:res_vits16} show that CDG-MAE substantially outperforms CropMAE across all downstream tasks, indicating the effectiveness of using diffusion generated views over crops of an image. CDG-MAE-a3 outperforms the baseline MAE ViT-B/16 (the backbone used to compute the metrics in Table~\ref{tab:metrics} and Table~\ref{tab:choice_of_diffusion_model}) by up to 10\%. Furthermore, CDG-MAE-a3 achieves better performance than SiamMAE across most metrics. This is noteworthy because CDG-MAE is pretrained on the ImageNet-1K dataset, whereas SiamMAE is trained with video frames from  Kinetics-400 dataset, and downstream evaluation tasks are video-based. This finding demonstrates the efficacy of our multi-anchor and anchor masking strategy for learning correspondences from static images.

\begin{figure}
    \centering
    \includegraphics[width=0.9\linewidth]{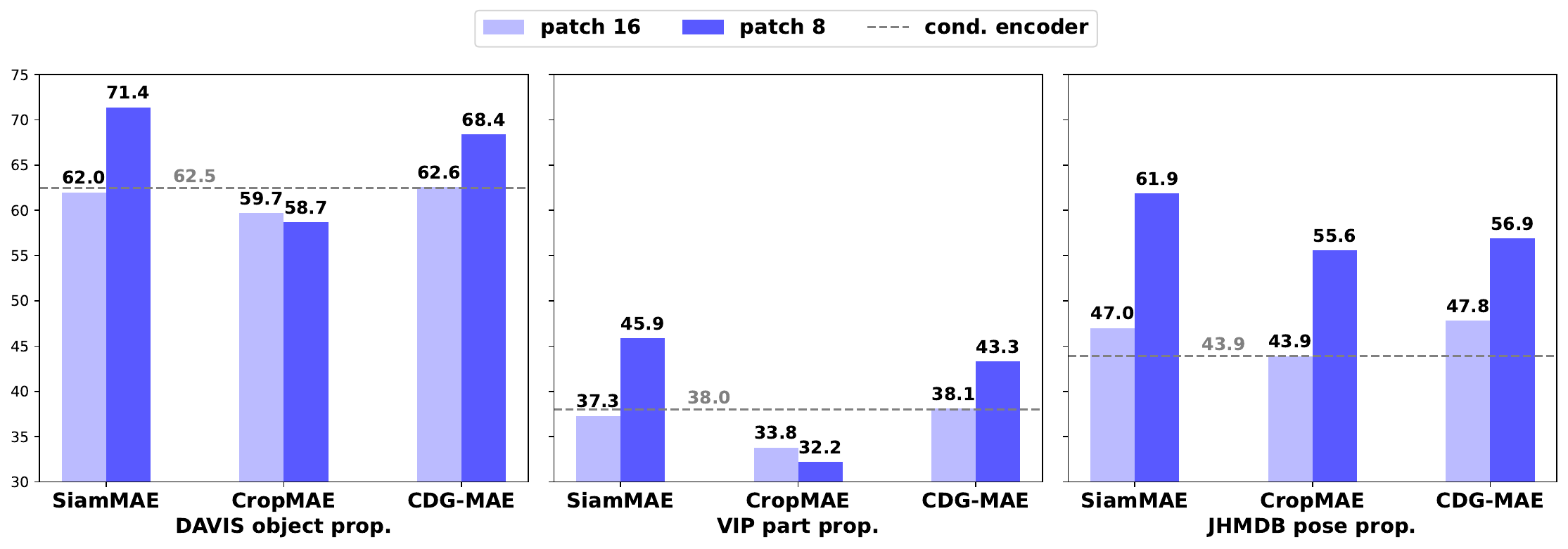}
    \caption{Performance of SiamMAE, CropMAE and CDG-MAE with ViT-S/16 and ViT-S/8 on DAVIS ($\mathcal{J} \& \mathcal{F}_m$), VIP (mIoU), and JHMDB (PCK\@0.1). We also present the performance of S-LDM's conditioning encoder with a dashed line.}
    \label{fig:res_patch8}
\end{figure}

\subsection{Impact of finer grained representations}
\label{sec:res_scaling}

SiamMAE has demonstrated that reducing the encoder patch size from $16$ to $8$ can substantially enhance performance. At patch size 8, SiamMAE can learn more robust correspondences from fine-grained changes (variations of small objects) between two frames. We investigate the effect of smaller patch size for CDG-MAE, and compare with SiamMAE and CropMAE by training a ViT-S/8 encoder. Results are presented in Figure~\ref{fig:res_patch8}.

CropMAE shows no performance improvement with ViT-S/8 in object and part propagation. This is likely because cropped views lack sufficient fine-grained variations to benefit smaller patch training. In contrast, both SiamMAE and CDG-MAE exhibit notable performance gains with ViT-S/8 across all three tasks, compared to their ViT-S/16 variants. CDG-MAE substantially outperforms image-based CropMAE and narrows the gap to the video-based SiamMAE. This highlights the ability of diffusion-generated views to provide diverse and rich variations beneficial for correspondence learning.


Furthermore, we evaluate the conditioning encoder of S-LDM ~\citep{gen-sis} on these downstream tasks. As indicated by the dashed line in Figure~\ref{fig:res_patch8}, while this encoder performs adequately on object and part propagation when compared to our ViT-S/16 encoder, it substantially underperforms in pose propagation. CDG-MAE ViT-S/8 outperforms the conditioning encoder in all three tasks. This shows that scaling tokens (patch size reduction) allows to better leverage diffusion-generated data, outperforming the original encoder used to condition the generation of such data.

\section{Storage and compute}
\label{sec:compute_storage}
Table~\ref{tab:compute_storage} details the storage requirements, GPU hours, and GFLOPs for SiamMAE, CropMAE, and our CDG-MAE variants. Specifically, CDG-MAE-a1 denotes the configuration with a single unmasked anchor; CDG-MAE-a2 uses two anchors with a 50\% anchor masking ratio; and CDG-MAE-a3 employs three anchors with a 25\% anchor masking ratio. To construct our bag of views, we generate $M=4$ synthetic views for each training image in ImageNet-1K (1.28M images) and store them on disk. This offline bag of views generation is a one-time preprocessing step that requires approximately 48 hours on a single node equipped with eight NVIDIA A100 GPUs. The total disk utilization for this synthetic dataset is 84~GB (amounting to 21~GB per full set of generated views). In comparison, the original ImageNet-1K dataset requires 140~GB. The synthetic views occupy less storage because they are saved at a fixed resolution of $256 \times 256$ pixels (the native output size of the diffusion model), whereas the original ImageNet images feature varying resolutions that are frequently larger than $256 \times 256$.

All pretraining experiments are conducted on a single node equipped with eight NVIDIA A100 (40 GB) GPUs. The GFLOPs for CropMAE and SiamMAE are marginally lower than CDG-MAE-a1 due to differences in their target masking ratios (98.5\% for CropMAE and 95\% for SiamMAE, compared to 90\% for CDG-MAE). Consequently, CropMAE and CDG-MAE-a1 require approximately the same pretraining time. Because the official code and weights for SiamMAE are not publicly available, we estimate its computational footprint by mapping its 2000-epoch training on Kinetics to an equivalent 374 epochs on ImageNet-1K (matching the total number of training iterations) . While extending CropMAE's training to 400 epochs (36 GPU hours) improves performance on JHMDB, it degrades performance on DAVIS and VIP; crucially, it still underperforms CDG-MAE-a1 trained for only 100 epochs. Refer to Appendix~\ref{sec:sup_cropmae400} for a more detailed comparison. We are unable to experiment with extended training epochs for SiamMAE because its official source code is not publicly available. Furthermore, despite utilizing two anchors, CDG-MAE-a2 requires similar pretraining hours to CDG-MAE-a1 due to our anchor masking strategy, while delivering superior performance. Finally, the offline bag of views is a one-time data preparation cost (requiring 84 GB of storage and 48 GPU hours) shared across all model variants. We believe that as diffusion models become increasingly fast and efficient, this preprocessing overhead will significantly decrease in the future. 



\begin{table}[!htbp]
\centering
{
\caption{Comparison of storage, GPU hours, and GFLOPs among SiamMAE, CropMAE, and CDG-MAE. The notations a1, a2, and a3 denote the single-anchor setting, the two-anchor setting with 50\% anchor masking, and the three-anchor setting with 25\% anchor masking, respectively. The offline bag of views generation is a one-time data preparation cost shared across all CDG-MAE model variants.}

\label{tab:compute_storage}
\resizebox{\columnwidth}{!}{
\begin{tabular}{c|ccc|cc|cc|c|c|c|c}
\toprule
   Method & Arch & Dataset & Epochs & \multicolumn{2}{c|} {Data Storage (GB)} &\multicolumn{2}{c|} {GPU hours} & GFLOPS & \multicolumn{1}{c|} {DAVIS} & VIP & \multicolumn{1}{c}{JHMDB} \\
    & & & & Original & Synthetic Views & Offline Gen. & Pretraining & & $\mathcal{J} \& \mathcal{F}_m$ & mIoU & PCK\@0.1 \\
    \midrule 
    SiamMAE &  ViT-S/16 & Kinetics-400 & 2000 & 348  & -- & -- & 34 & 5.8 & 62.0 & 37.3 & 47.0   \\
    \midrule
    CropMAE &  ViT-S/16 & ImageNet-1K & 100 &  140  & -- & -- & 9 & 5.7 & 59.7 & 33.8 & 43.9   \\
    CropMAE &  ViT-S/16 & ImageNet-1K & 400 &  140 & -- & -- & 36&  5.7 & 55.5 & 30.7 & 44.3   \\
    \midrule
    \rowcolor{blue!10}
    CDG-MAE-a1 &  ViT-S/16 & ImageNet-1K & 100 & 140  & 84 & 48 & 9 & 6.0 & 61.2 & 37.6 & 46.5   \\
    CDG-MAE-a2 &  ViT-S/16 & ImageNet-1K & 100 & 140  & 84 & 48 & 9 & 6.1 & 62.1 & 38.1 & 47.8   \\
    \rowcolor{blue!10}
    CDG-MAE-a3 &  ViT-S/16 & ImageNet-1K & 100 & 140  & 84 & 48 & 14 & 11.6 & 62.6 & 38.1 & 47.8   \\
    
\end{tabular}}}
\end{table}


\section{Conclusion}
 \label{sec:conclusion}  
We introduced CDG-MAE, a novel MAE framework for cross-view pretraining using diffusion-generated views. We developed new metrics to evaluate the local and global consistency of generated views. Such properties, inherent in video data, are important to learn correspondences. We demonstrate the effectiveness of the proposed consistency metrics in choosing the right diffusion model for view generation. CDG-MAE, trained with diffusion views derived from static images along with our proposed multi-anchor masking, substantially outperforms existing crop-based MAE methods and narrows the performance gap with video-based approaches. We hope our work inspires further exploration into synthetic data generation to leverage rich and diverse image datasets for cross-view representation learning. 

\myparagraph{Limitations and future work} Although self-supervised diffusion models can generate diverse variations needed for cross-view pretraining, we cannot control which specific variations occur between generated views. Future work can study how to better control variations, e.g. pose changes, in the generated views. This direction is challenging and interesting to explore as this should be done in a self-supervised way, i.e. without using pose labels.

\myparagraph{Broader impact} Our models inherit the biases present in the public datasets used for training. The synthetic data generation from diffusion models can further amplify the bias present in the pretraining datasets. As noted in Section~\ref{sec:compute_storage}, large-scale view generation requires approximately 48 hours. Scaling this approach to larger diffusion backbones or massive datasets beyond ImageNet involves higher computational overhead and energy consumption. We expect these requirements to decrease as diffusion models become more computationally efficient. Finally, utilizing our framework for pretraining within safety-critical pipelines demands careful consideration. It may require strict human verification and quality filtering of the generated views to safeguard against the systematic risks associated with generative models.

\section{Acknowledgments}

This research has been supported by DATAIA international mobility scholarship,  by the project MIS 5154714 of the National Recovery and
Resilience Plan Greece 2.0 funded by the European Union under the NextGenerationEU Program, NCI awards
1R21CA258493-01A1, 5U24CA215109, UH3CA225021,
U24CA180924, NSF grants IIS-2123920, IIS-2212046, Stony Brook Profund 2022 seed funding. This research used resources of the Argonne Leadership Computing Facility, a U.S. Department of Energy (DOE) Office of Science user facility at Argonne National Laboratory and is based on research supported by the U.S. DOE Office of Science-Advanced Scientific Computing Research Program, under Contract No. DE-AC02-06CH11357. Furthermore, this work has been supported by the Agence Nationale de la Recherche through ANR-23-IAHU-0002, ANR-21-CE45-0007, ANR-23-CE45-0029, ANR-23-IACL-0003 (DATAIA CLUSTER) and the Health Data Hub as part of the second edition of the France-Québec call for projects Intelligence Artificielle en santé.
\clearpage

\bibliography{main}

@article{weinzaepfel2022croco,
  title={Croco: Self-supervised pre-training for 3d vision tasks by cross-view completion},
  author={Weinzaepfel, Philippe and Leroy, Vincent and Lucas, Thomas and Br{\'e}gier, Romain and Cabon, Yohann and Arora, Vaibhav and Antsfeld, Leonid and Chidlovskii, Boris and Csurka, Gabriela and Revaud, J{\'e}r{\^o}me},
  journal={NeurIPS},
  year={2022}
}

@inproceedings{weinzaepfel2023croco,
  title={Croco v2: Improved cross-view completion pre-training for stereo matching and optical flow},
  author={Weinzaepfel, Philippe and Lucas, Thomas and Leroy, Vincent and Cabon, Yohann and Arora, Vaibhav and Br{\'e}gier, Romain and Csurka, Gabriela and Antsfeld, Leonid and Chidlovskii, Boris and Revaud, J{\'e}r{\^o}me},
  booktitle={ICCV},
  year={2023}
}

@article{gupta2023siamese,
  title={Siamese masked autoencoders},
  author={Gupta, Agrim and Wu, Jiajun and Deng, Jia and Li, Fei-Fei},
  journal={NeurIPS},
  year={2023}
}

@inproceedings{eymael2024efficient,
  title={Efficient image pre-training with siamese cropped masked autoencoders},
  author={Eyma{\"e}l, Alexandre and Vandeghen, Renaud and Cioppa, Anthony and Giancola, Silvio and Ghanem, Bernard and Van Droogenbroeck, Marc},
  booktitle={ECCV},
  year={2024},
}

@inproceedings{rombach2022high,
  title={High-resolution image synthesis with latent diffusion models},
  author={Rombach, Robin and Blattmann, Andreas and Lorenz, Dominik and Esser, Patrick and Ommer, Bj{\"o}rn},
  booktitle={Proceedings of the IEEE/CVF conference on computer vision and pattern recognition},
  pages={10684--10695},
  year={2022}
}

@article{dhariwal2021diffusion,
  title={Diffusion models beat gans on image synthesis},
  author={Dhariwal, Prafulla and Nichol, Alexander},
  journal={Advances in neural information processing systems},
  volume={34},
  pages={8780--8794},
  year={2021}
}

@misc{flux2024,
    author={Black Forest Labs},
    title={FLUX},
    year={2024},
    howpublished={\url{https://github.com/black-forest-labs/flux}},
}

@inproceedings{zhang2023adding,
  title={Adding conditional control to text-to-image diffusion models},
  author={Zhang, Lvmin and Rao, Anyi and Agrawala, Maneesh},
  booktitle={Proceedings of the IEEE/CVF international conference on computer vision},
  pages={3836--3847},
  year={2023}
}

@InProceedings{Graikos_2024_CVPR,
    author    = {Graikos, Alexandros and Yellapragada, Srikar and Le, Minh-Quan and Kapse, Saarthak and Prasanna, Prateek and Saltz, Joel and Samaras, Dimitris},
    title     = {Learned Representation-Guided Diffusion Models for Large-Image Generation},
    booktitle = {Proceedings of the IEEE/CVF Conference on Computer Vision and Pattern Recognition (CVPR)},
    month     = {June},
    year      = {2024},
    pages     = {8532-8542}
}

@misc{gen-sis,
      title={Gen-SIS: Generative Self-augmentation Improves Self-supervised Learning}, 
      author={Varun Belagali and Srikar Yellapragada and Alexandros Graikos and Saarthak Kapse and Zilinghan Li and Tarak Nath Nandi and Ravi K Madduri and Prateek Prasanna and Joel Saltz and Dimitris Samaras},
      year={2024},
      eprint={2412.01672},
      archivePrefix={arXiv},
      primaryClass={cs.CV},
      url={https://arxiv.org/abs/2412.01672}, 
}

@article{davis-vos,
  title={The 2017 davis challenge on video object segmentation},
  author={Pont-Tuset, Jordi and Perazzi, Federico and Caelles, Sergi and Arbel{\'a}ez, Pablo and Sorkine-Hornung, Alex and Van Gool, Luc},
  journal={arXiv preprint arXiv:1704.00675},
  year={2017}
}

@inproceedings{jhmdb,
  title={Towards understanding action recognition},
  author={Jhuang, Hueihan and Gall, Juergen and Zuffi, Silvia and Schmid, Cordelia and Black, Michael J},
  booktitle={Proceedings of the IEEE international conference on computer vision},
  pages={3192--3199},
  year={2013}
}

@inproceedings{vip,
  title={Adaptive temporal encoding network for video instance-level human parsing},
  author={Zhou, Qixian and Liang, Xiaodan and Gong, Ke and Lin, Liang},
  booktitle={Proceedings of the 26th ACM international conference on Multimedia},
  pages={1527--1535},
  year={2018}
}

@article{bao2021beit,
  title={Beit: Bert pre-training of image transformers},
  author={Bao, Hangbo and Dong, Li and Piao, Songhao and Wei, Furu},
  journal={arXiv preprint arXiv:2106.08254},
  year={2021}
}

@article{tong2022videomae,
  title={Videomae: Masked autoencoders are data-efficient learners for self-supervised video pre-training},
  author={Tong, Zhan and Song, Yibing and Wang, Jue and Wang, Limin},
  journal={Advances in neural information processing systems},
  volume={35},
  pages={10078--10093},
  year={2022}
}

@inproceedings{xie2022simmim,
  title={Simmim: A simple framework for masked image modeling},
  author={Xie, Zhenda and Zhang, Zheng and Cao, Yue and Lin, Yutong and Bao, Jianmin and Yao, Zhuliang and Dai, Qi and Hu, Han},
  booktitle={Proceedings of the IEEE/CVF conference on computer vision and pattern recognition},
  pages={9653--9663},
  year={2022}
}

@inproceedings{wei2022masked,
  title={Masked feature prediction for self-supervised visual pre-training},
  author={Wei, Chen and Fan, Haoqi and Xie, Saining and Wu, Chao-Yuan and Yuille, Alan and Feichtenhofer, Christoph},
  booktitle={Proceedings of the IEEE/CVF conference on computer vision and pattern recognition},
  pages={14668--14678},
  year={2022}
}

@inproceedings{he2022masked,
  title={Masked autoencoders are scalable vision learners},
  author={He, Kaiming and Chen, Xinlei and Xie, Saining and Li, Yanghao and Doll{\'a}r, Piotr and Girshick, Ross},
  booktitle={Proceedings of the IEEE/CVF conference on computer vision and pattern recognition},
  pages={16000--16009},
  year={2022}
}

@inproceedings{jepa,
  title={Self-supervised learning from images with a joint-embedding predictive architecture},
  author={Assran, Mahmoud and Duval, Quentin and Misra, Ishan and Bojanowski, Piotr and Vincent, Pascal and Rabbat, Michael and LeCun, Yann and Ballas, Nicolas},
  booktitle={Proceedings of the IEEE/CVF Conference on Computer Vision and Pattern Recognition},
  pages={15619--15629},
  year={2023}
}

@article{ho2022classifier,
  title={Classifier-free diffusion guidance},
  author={Ho, Jonathan and Salimans, Tim},
  journal={arXiv preprint arXiv:2207.12598},
  year={2022}
}

@article{song2020denoising,
  title={Denoising diffusion implicit models},
  author={Song, Jiaming and Meng, Chenlin and Ermon, Stefano},
  journal={arXiv preprint arXiv:2010.02502},
  year={2020}
}

@article{rcg,
  title={Return of unconditional generation: A self-supervised representation generation method},
  author={Li, Tianhong and Katabi, Dina and He, Kaiming},
  journal={Advances in Neural Information Processing Systems},
  volume={37},
  pages={125441--125468},
  year={2024}
}

@inproceedings{dinov1,
  title={Emerging properties in self-supervised vision transformers},
  author={Caron, Mathilde and Touvron, Hugo and Misra, Ishan and J{\'e}gou, Herv{\'e} and Mairal, Julien and Bojanowski, Piotr and Joulin, Armand},
  booktitle={Proceedings of the IEEE/CVF international conference on computer vision},
  pages={9650--9660},
  year={2021}
}

@inproceedings{moco,
  title={An empirical study of training self-supervised vision transformers},
  author={Chen, Xinlei and Xie, Saining and He, Kaiming},
  booktitle={Proceedings of the IEEE/CVF international conference on computer vision},
  pages={9640--9649},
  year={2021}
}

@inproceedings{peebles2023scalable,
  title={Scalable diffusion models with transformers},
  author={Peebles, William and Xie, Saining},
  booktitle={Proceedings of the IEEE/CVF international conference on computer vision},
  pages={4195--4205},
  year={2023}
}

@misc{esser2024scalingrectifiedflowtransformers,
      title={Scaling Rectified Flow Transformers for High-Resolution Image Synthesis}, 
      author={Patrick Esser and Sumith Kulal and Andreas Blattmann and Rahim Entezari and Jonas Müller and Harry Saini and Yam Levi and Dominik Lorenz and Axel Sauer and Frederic Boesel and Dustin Podell and Tim Dockhorn and Zion English and Kyle Lacey and Alex Goodwin and Yannik Marek and Robin Rombach},
      year={2024},
      eprint={2403.03206},
      archivePrefix={arXiv},
      primaryClass={cs.CV},
      url={https://arxiv.org/abs/2403.03206}, 
}

@misc{lu2023dpmsolverfastsolverguided,
      title={DPM-Solver++: Fast Solver for Guided Sampling of Diffusion Probabilistic Models}, 
      author={Cheng Lu and Yuhao Zhou and Fan Bao and Jianfei Chen and Chongxuan Li and Jun Zhu},
      year={2023},
      eprint={2211.01095},
      archivePrefix={arXiv},
      primaryClass={cs.LG},
      url={https://arxiv.org/abs/2211.01095}, 
}

@article{tian2024stablerep,
  title={Stablerep: Synthetic images from text-to-image models make strong visual representation learners},
  author={Tian, Yonglong and Fan, Lijie and Isola, Phillip and Chang, Huiwen and Krishnan, Dilip},
  journal={Advances in Neural Information Processing Systems},
  volume={36},
  year={2024}
}

@inproceedings{tian2024learning,
  title={Learning vision from models rivals learning vision from data},
  author={Tian, Yonglong and Fan, Lijie and Chen, Kaifeng and Katabi, Dina and Krishnan, Dilip and Isola, Phillip},
  booktitle={Proceedings of the IEEE/CVF conference on computer vision and pattern recognition},
  pages={15887--15898},
  year={2024}
}

@inproceedings{chen2020simple,
  title={A simple framework for contrastive learning of visual representations},
  author={Chen, Ting and Kornblith, Simon and Norouzi, Mohammad and Hinton, Geoffrey},
  booktitle={International conference on machine learning},
  pages={1597--1607},
  year={2020},
  organization={PmLR}
}

@article{grill2020bootstrap,
  title={Bootstrap your own latent-a new approach to self-supervised learning},
  author={Grill, Jean-Bastien and Strub, Florian and Altch{\'e}, Florent and Tallec, Corentin and Richemond, Pierre and Buchatskaya, Elena and Doersch, Carl and Avila Pires, Bernardo and Guo, Zhaohan and Gheshlaghi Azar, Mohammad and others},
  journal={Advances in neural information processing systems},
  volume={33},
  pages={21271--21284},
  year={2020}
}

@inproceedings{he2020momentum,
  title={Momentum contrast for unsupervised visual representation learning},
  author={He, Kaiming and Fan, Haoqi and Wu, Yuxin and Xie, Saining and Girshick, Ross},
  booktitle={Proceedings of the IEEE/CVF conference on computer vision and pattern recognition},
  pages={9729--9738},
  year={2020}
}

@article{zhou2021ibot,
  title={ibot: Image bert pre-training with online tokenizer},
  author={Zhou, Jinghao and Wei, Chen and Wang, Huiyu and Shen, Wei and Xie, Cihang and Yuille, Alan and Kong, Tao},
  journal={arXiv preprint arXiv:2111.07832},
  year={2021}
}

@article{oquab2023dinov2,
  title={Dinov2: Learning robust visual features without supervision},
  author={Oquab, Maxime and Darcet, Timoth{\'e}e and Moutakanni, Th{\'e}o and Vo, Huy and Szafraniec, Marc and Khalidov, Vasil and Fernandez, Pierre and Haziza, Daniel and Massa, Francisco and El-Nouby, Alaaeldin and others},
  journal={arXiv preprint arXiv:2304.07193},
  year={2023}
}

@article{bardes2022vicregl,
  title={Vicregl: Self-supervised learning of local visual features},
  author={Bardes, Adrien and Ponce, Jean and LeCun, Yann},
  journal={Advances in Neural Information Processing Systems},
  volume={35},
  pages={8799--8810},
  year={2022}
}

@article{feichtenhofer2022masked,
  title={Masked autoencoders as spatiotemporal learners},
  author={Feichtenhofer, Christoph and Li, Yanghao and He, Kaiming and others},
  journal={Advances in neural information processing systems},
  volume={35},
  pages={35946--35958},
  year={2022}
}

@inproceedings{deng2009imagenet,
  title={Imagenet: A large-scale hierarchical image database},
  author={Deng, Jia and Dong, Wei and Socher, Richard and Li, Li-Jia and Li, Kai and Fei-Fei, Li},
  booktitle={2009 IEEE conference on computer vision and pattern recognition},
  pages={248--255},
  year={2009},
  organization={Ieee}
}

@article{kay2017kinetics,
  title={The kinetics human action video dataset},
  author={Kay, Will and Carreira, Joao and Simonyan, Karen and Zhang, Brian and Hillier, Chloe and Vijayanarasimhan, Sudheendra and Viola, Fabio and Green, Tim and Back, Trevor and Natsev, Paul and others},
  journal={arXiv preprint arXiv:1705.06950},
  year={2017}
}

@inproceedings{ma2025learning,
  title={Learning visual generative priors without text},
  author={Ma, Shuailei and Zheng, Kecheng and Wei, Ying and Wu, Wei and Lu, Fan and Zhang, Yifei and Xie, Chen-Wei and Gong, Biao and Zhu, Jiapeng and Shen, Yujun},
  booktitle={Proceedings of the Computer Vision and Pattern Recognition Conference},
  pages={8051--8061},
  year={2025}
}

@article{wan2025wan,
  title={Wan: Open and advanced large-scale video generative models},
  author={Wan, Team and Wang, Ang and Ai, Baole and Wen, Bin and Mao, Chaojie and Xie, Chen-Wei and Chen, Di and Yu, Feiwu and Zhao, Haiming and Yang, Jianxiao and others},
  journal={arXiv preprint arXiv:2503.20314},
  year={2025}
}

@article{kong2024hunyuanvideo,
  title={Hunyuanvideo: A systematic framework for large video generative models},
  author={Kong, Weijie and Tian, Qi and Zhang, Zijian and Min, Rox and Dai, Zuozhuo and Zhou, Jin and Xiong, Jiangfeng and Li, Xin and Wu, Bo and Zhang, Jianwei and others},
  journal={arXiv preprint arXiv:2412.03603},
  year={2024}
}

@article{hacohen2024ltx,
  title={Ltx-video: Realtime video latent diffusion},
  author={HaCohen, Yoav and Chiprut, Nisan and Brazowski, Benny and Shalem, Daniel and Moshe, Dudu and Richardson, Eitan and Levin, Eran and Shiran, Guy and Zabari, Nir and Gordon, Ori and others},
  journal={arXiv preprint arXiv:2501.00103},
  year={2024}
}

@article{wang2025waft,
  title={Waft: Warping-alone field transforms for optical flow},
  author={Wang, Yihan and Deng, Jia},
  journal={arXiv preprint arXiv:2506.21526},
  year={2025}
}

@inproceedings{flyingchairs,
  title={Flownet: Learning optical flow with convolutional networks},
  author={Dosovitskiy, Alexey and Fischer, Philipp and Ilg, Eddy and Hausser, Philip and Hazirbas, Caner and Golkov, Vladimir and Van Der Smagt, Patrick and Cremers, Daniel and Brox, Thomas},
  booktitle={Proceedings of the IEEE international conference on computer vision},
  pages={2758--2766},
  year={2015}
}

@inproceedings{flyingthings,
  title={A large dataset to train convolutional networks for disparity, optical flow, and scene flow estimation},
  author={Mayer, Nikolaus and Ilg, Eddy and Hausser, Philip and Fischer, Philipp and Cremers, Daniel and Dosovitskiy, Alexey and Brox, Thomas},
  booktitle={Proceedings of the IEEE conference on computer vision and pattern recognition},
  pages={4040--4048},
  year={2016}
}

@article{kitti,
  title={Vision meets robotics: The kitti dataset},
  author={Geiger, Andreas and Lenz, Philip and Stiller, Christoph and Urtasun, Raquel},
  journal={The international journal of robotics research},
  volume={32},
  number={11},
  pages={1231--1237},
  year={2013},
  publisher={Sage Publications Sage UK: London, England}
}

@inproceedings{ranftl2021vision,
  title={Vision transformers for dense prediction},
  author={Ranftl, Ren{\'e} and Bochkovskiy, Alexey and Koltun, Vladlen},
  booktitle={Proceedings of the IEEE/CVF international conference on computer vision},
  pages={12179--12188},
  year={2021}
}

@article{cherti2022reproducible,
  title={Reproducible scaling laws for contrastive language-image learning},
  author={Cherti, Mehdi and Beaumont, Romain and Wightman, Ross and Wortsman, Mitchell and Ilharco, Gabriel and Gordon, Cade and Schuhmann, Christoph and Schmidt, Ludwig and Jitsev, Jenia},
  journal={arXiv preprint arXiv:2212.07143},
  year={2022}
}
\bibliographystyle{tmlr}

\newpage

\appendix
\section{Appendix}
\subsection{Implementation details}
\label{sec:sup_implementation}
\subsubsection{Training}
\label{sec:sup_training_details}
Our implementation is built using the CropMAE~\citep{eymael2024efficient} codebase. We use ImageNet-1K as our pretraining dataset. The default training hyperparameters for our method (CDG-MAE) under the single anchor setting are presented in Table~\ref{tab:sup_hyperparamters}. We also provide the hyperparameters of CropMAE as a reference. CDG-MAE also uses cropping as an augmentation, following SiamMAE~\citep{gupta2023siamese}.

\begin{table}[!htbp]
\centering
{
\caption{Training hyperparameters for CDG-MAE and CropMAE~\citep{eymael2024efficient}}
\label{tab:sup_hyperparamters}
\begin{tabular}{c|c|c}
\toprule
   \textbf{} & CDG-MAE (ours) & CropMAE  \\

   \midrule
    Optimizer &  AdamW ($\beta_1$=0.9, $\beta_2$=0.95 )	 & AdamW   ($\beta_1$=0.9, $\beta_2$=0.95 )  \\
    Weight decay & 0.05 & 0.05 \\ 
    Base learning rate & 1.5 $\times$ 10$^{-4}$ & 1.5 $\times$ 10$^{-4}$ \\
    Target masking ratio & 90\% & 98.5\% \\
    $lr$ schedule & Cosine Decay & Cosine Decay \\
    Epochs & 100 & 100 \\
    Batch size & 2048 & 2048 \\
    Bag of views size ($M$) & 4 & -- \\
    Augmentations &  Crop [0.5, 1.0] &  LocalToGlobal Crop [(0.3, 6.0) (0.1, 1.0)] \\
    Aspect ratio & (0.75, 1.33) & (0.75, 1.33) \\
    
\end{tabular}}
\end{table}

\begin{figure}
    \centering
    \includegraphics[width=0.9\linewidth]{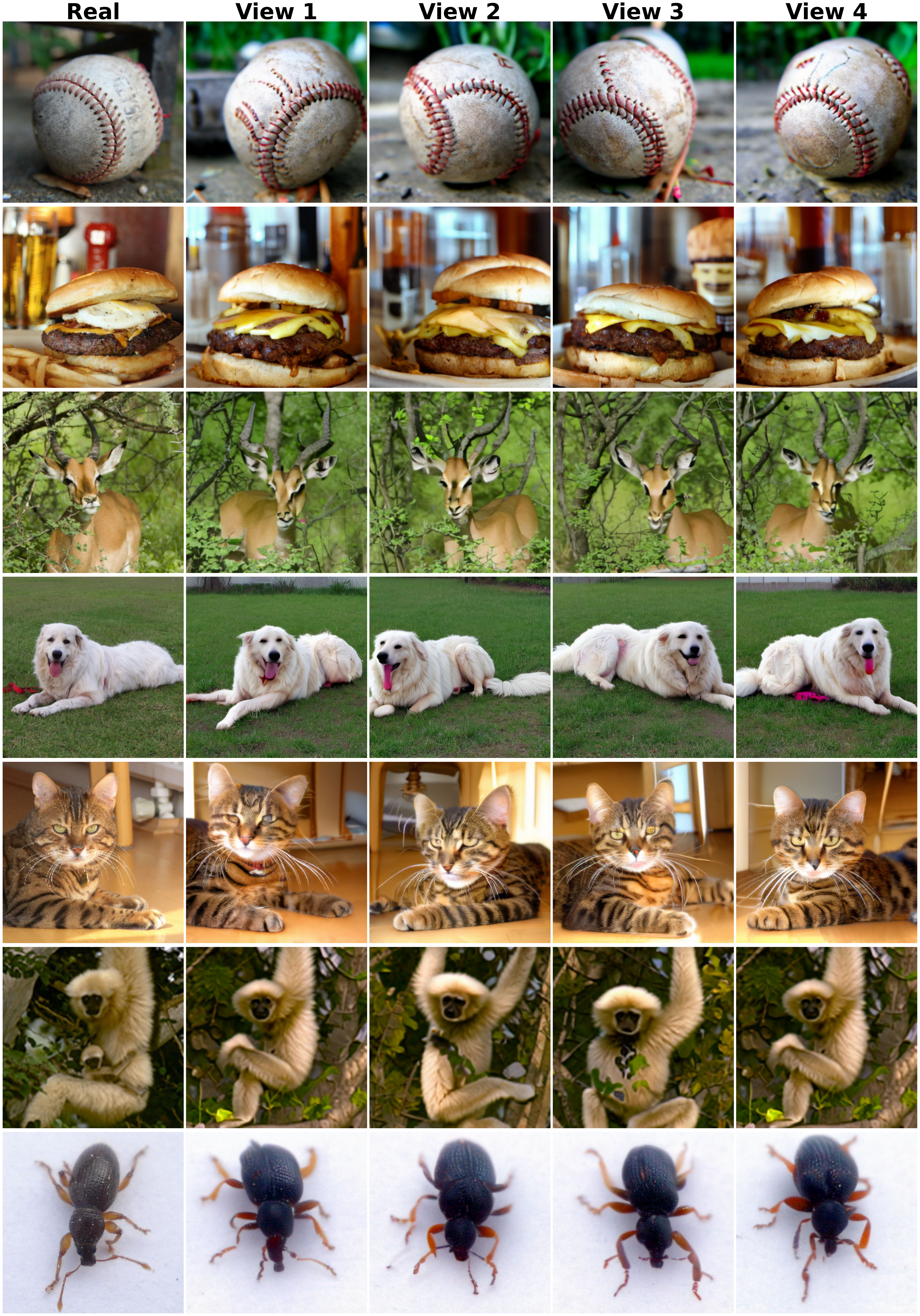}
    \caption{Bag of views visualization. Real denotes an image in ImageNet dataset. The views represent the synthetic views generated with diffusion.}
    \label{fig:sample_bag_views}
\end{figure}

\begin{figure}
    \centering
    \includegraphics[width=\linewidth]{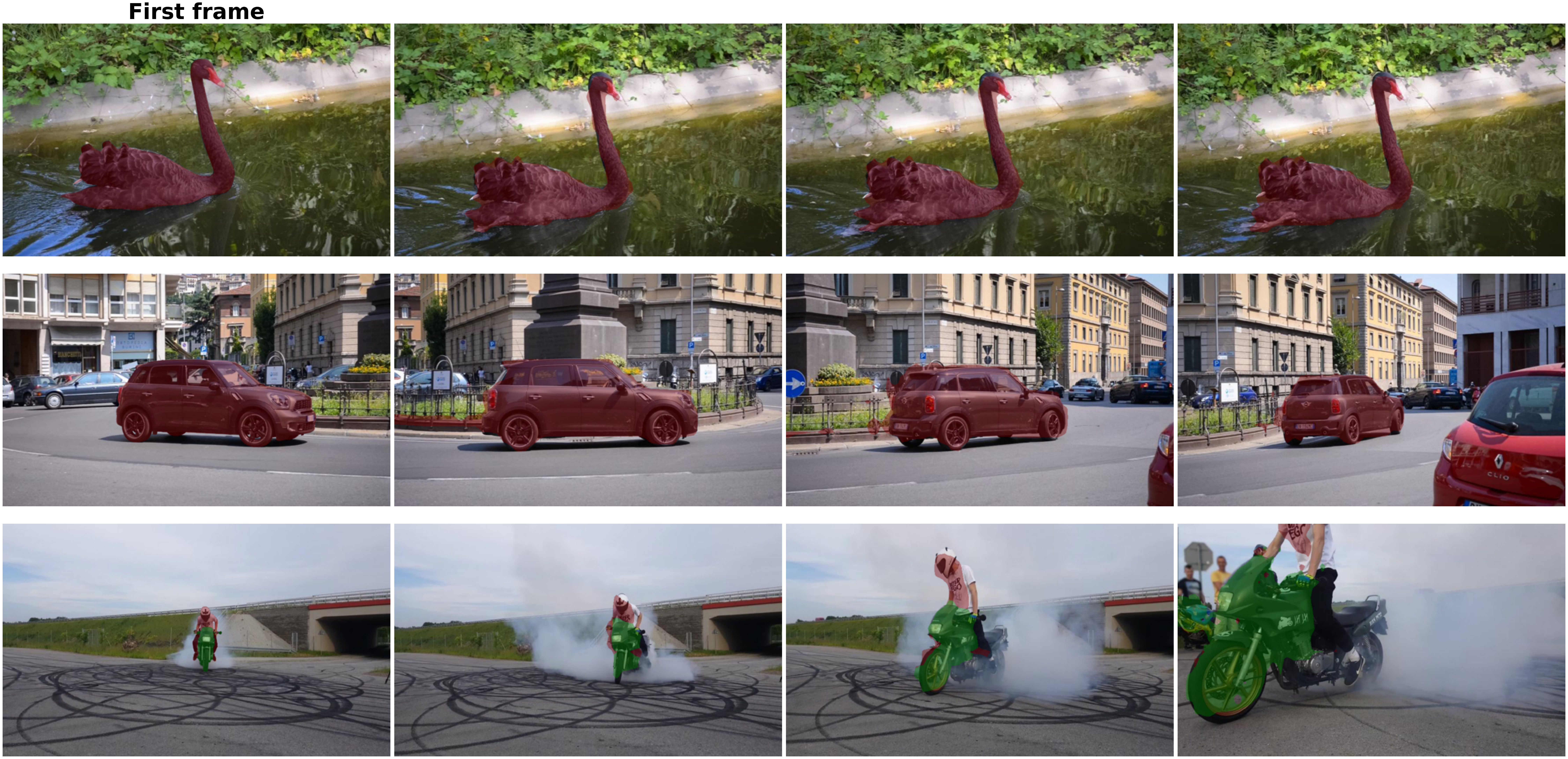}
    \caption{Visualization of label propagation using CDG-MAE ViT-S/16 on DAVIS~\cite{davis-vos} dataset. The first frame is annotated with the ground truth object segmentation masks.}
    \label{fig:sup_vis_davis}
\end{figure}

\begin{figure}
    \centering
    \includegraphics[width=\linewidth]{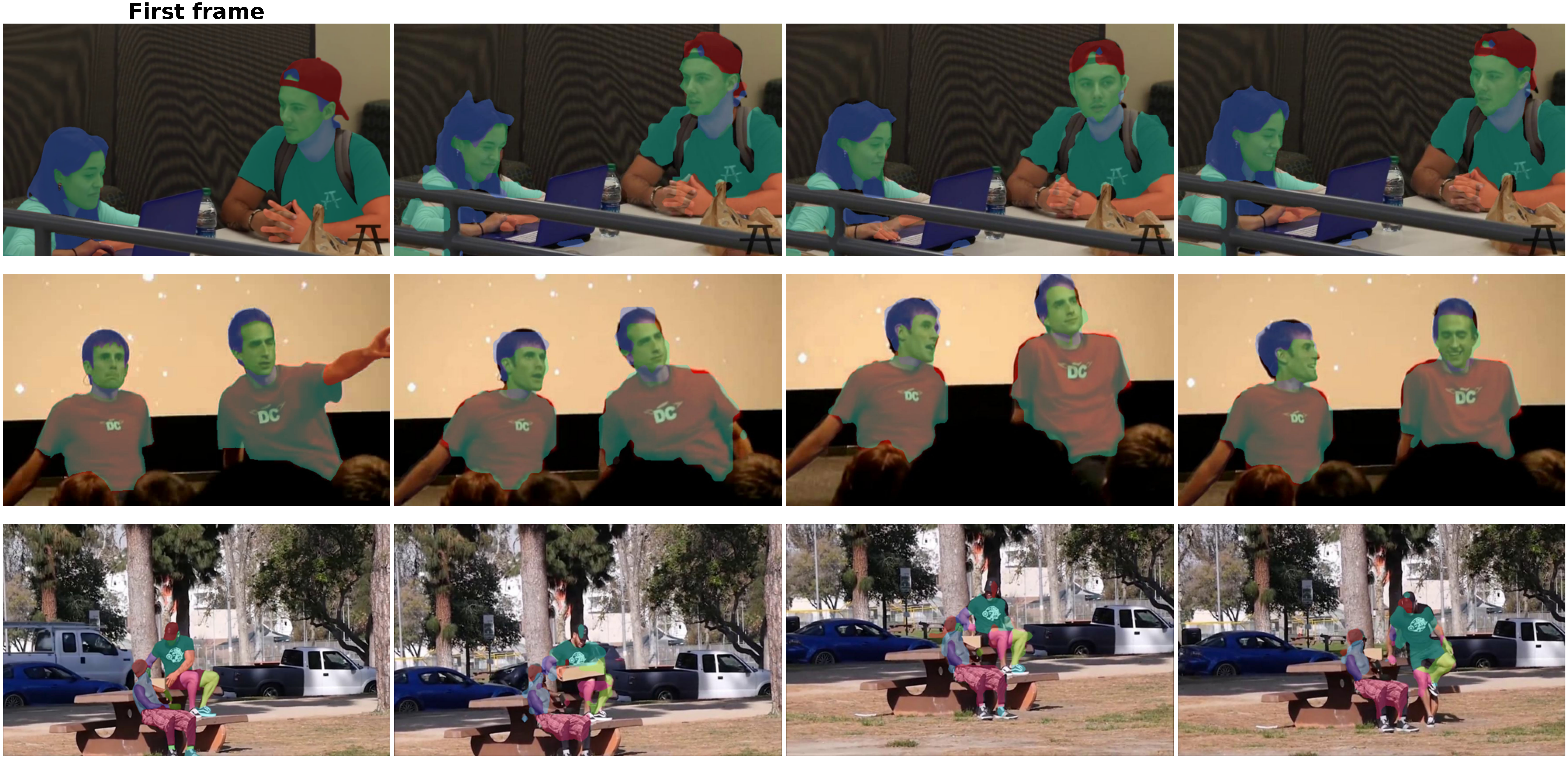}
    \caption{Visualization of label propagation using CDG-MAE ViT-S/16 on VIP~\cite{vip} dataset. The first frame is annotated with the semantic part segmentation masks.}
    \label{fig:sup_vis_vip}
\end{figure}

\begin{figure}
    \centering
    \includegraphics[width=\linewidth]{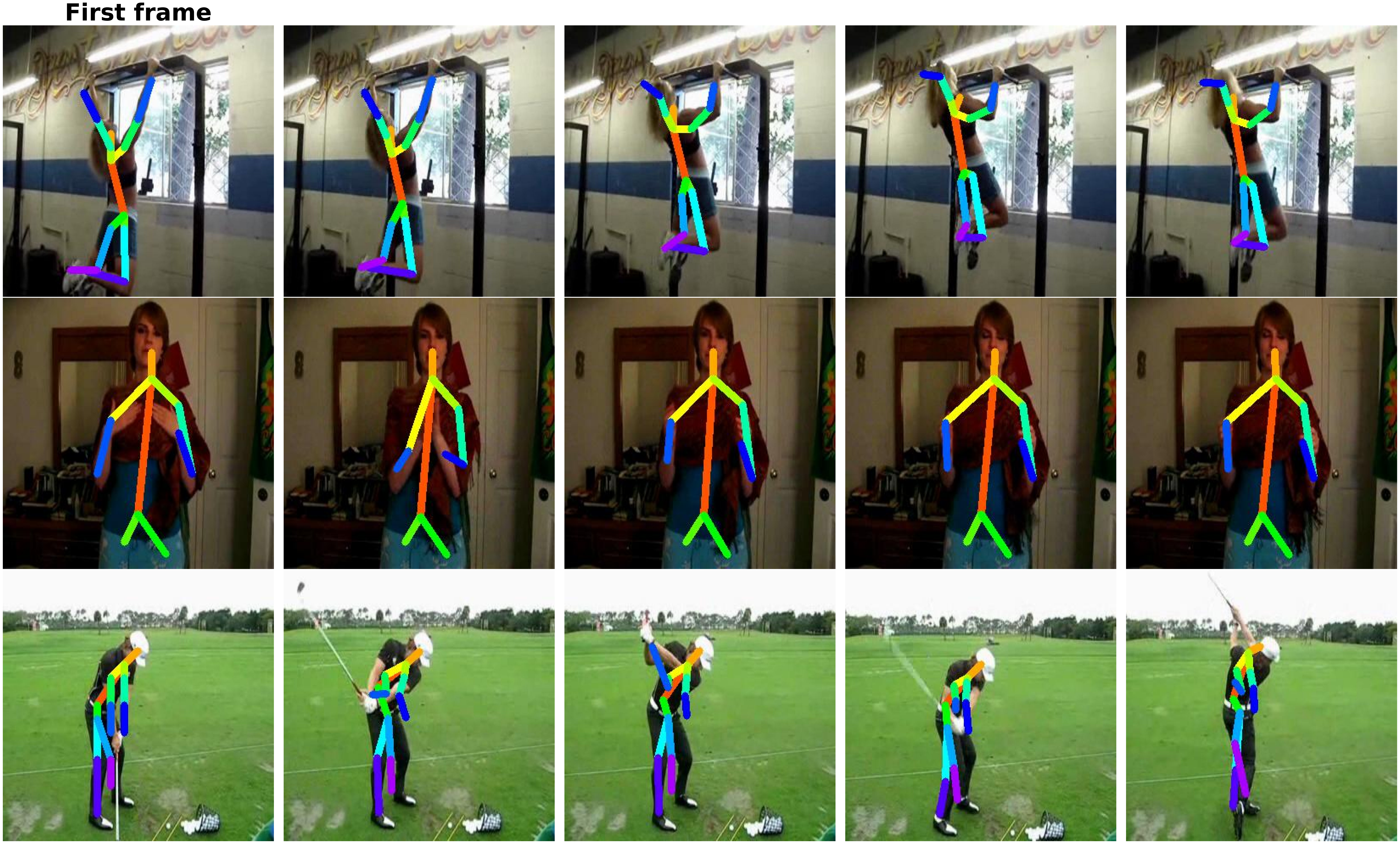}
    \caption{Visualization of label propagation using CDG-MAE ViT-S/16 on JHMDB~\cite{jhmdb} dataset. The first frame is annotated with the pose labels.}
    \label{fig:sup_vis_jhmdb}
\end{figure}

\subsubsection{Downstream Evaluation}
\label{sec:sup_down_eval}
We evaluate our method on standard video label propagation tasks, following the evaluation protocol of previous works~\citep{eymael2024efficient, gupta2023siamese}. We use three datasets: 1) DAVIS~\citep{davis-vos} for video object segmentation, 2) VIP~\citep{vip} for semantic part propagation, and 3) JHMDB~\citep{jhmdb} for human pose propagation. In these tasks, annotation is provided for the first frame, and the objective is to propagate ground-truth labels to all subsequent frames of the video.

The evaluation is performed in a training-free manner using $k$-nearest neighbor ($k$-NN) inference. Furthermore, this protocol utilizes a memory queue of the last few frames and restricts source patches to the query's spatial neighborhood. The specific hyperparameter values for this setup are detailed in Table~\ref{tab:sup_hyperparamters_down}.

For DAVIS, we report mean region similarity ($\mathcal{J}_m$), mean contour accuracy ($\mathcal{F}_m$), and their combined average ($\mathcal{J} \& \mathcal{F}_m$). For VIP, we report the mean Intersection over Union (mIoU). For JHMDB, evaluation is based on PCK0.1 and PCK0.2, which represent the percentage of keypoints correctly localized within an error margin of 10\% and 20\% of the bounding box size, respectively. We use the evaluation codebase released by CropMAE~\citep{eymael2024efficient}.

\begin{table}[!htbp]
\centering
{\small
\caption{Hyperparameters for downstream evaluation using $k$-nearest neighbor ($k$-NN) inference.}
\label{tab:sup_hyperparamters_down}
\begin{tabular}{c|c|c|c}
\toprule
   \textbf{} & DAVIS & VIP & JHMDB \\

   \midrule
    Top-K &  7	&   10 &  7 \\
    Queue Length & 20 & 20  & 20 \\ 
    Neighborhood  Size    & 20 & 20  & 20 \\

\end{tabular}}
\end{table}

\subsection{Visualization}
\label{sec:sup_vis}
Figure~\ref{fig:sample_bag_views} shows samples from the bag of views generated using the ImageNet-1K dataset. As seen in the figure, the views exhibit changes in pose, motion, and perspective. Moreover, one can observe that the generated images maintain the main characteristics of the image (objects and background) making them ideal for the training of our CDG-MAE method.

Qualitative results of CDG-MAE on downstream tasks are presented in Figure~\ref{fig:sup_vis_davis}, Figure~\ref{fig:sup_vis_vip}, and Figure~\ref{fig:sup_vis_jhmdb}.

\subsection{Machine details and total budget}
\label{sec:sup_compute_budget}
We use a single node with 8 A100 40 GB GPUs for all the experiments. The CDG-MAE ViT-S/16 model takes a maximum of 14 hours for training. A total of 900 node hours were used for this paper, including initial exploration and failed experiments.

\subsection{Additional experiments}
\label{sec:sup_experiments}

\subsubsection{Training CropMAE for more epochs}
\label{sec:sup_cropmae400}
\begin{figure}
    \centering
    \includegraphics[width=\linewidth]{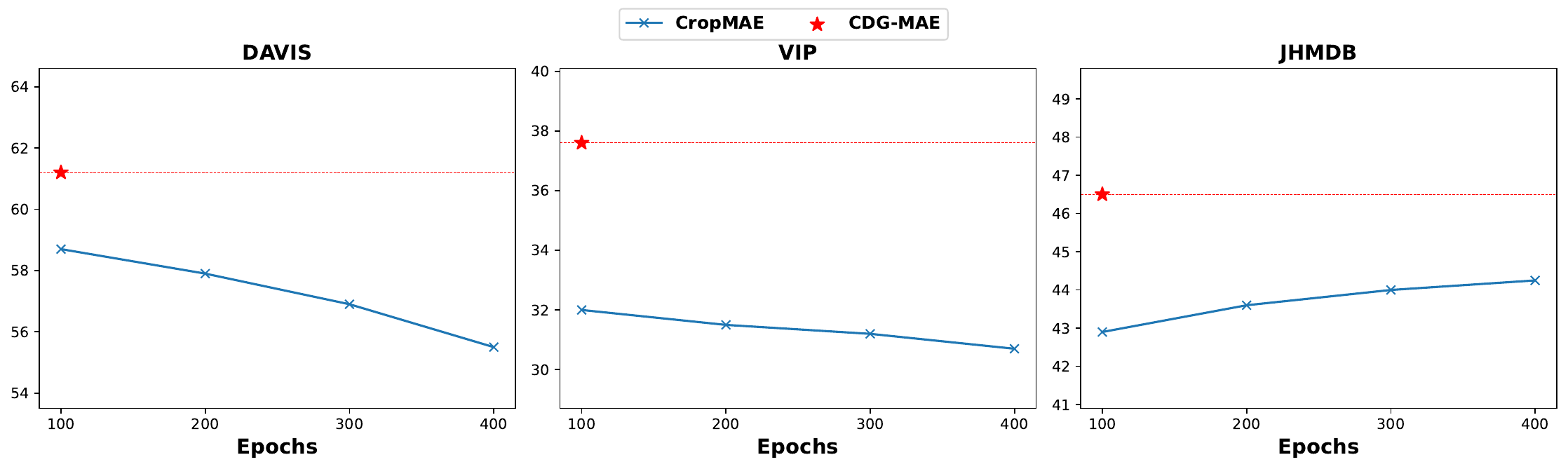}
    \caption{Performance of CropMAE when trained with 400 epochs schedule. We report the evaluation on DAVIS ($\mathcal{J} \& \mathcal{F}_m$), VIP (mIoU), and JHMDB (PCK\@0.1). We also present the performance of CDG-MAE trained for 100 epochs with single anchor setting.}
    \label{fig:sup_cropmae_ep400}
\end{figure}

We compare the performance of CDG-MAE trained for 100 epochs with CropMAE trained for a longer schedule (400 epochs). As studied by CropMAE~\citep{eymael2024efficient} paper, extended training might lead to saturation in performance. In Figure~\ref{fig:sup_cropmae_ep400}, we observe a similar trend, where the performance of CropMAE starts to decrease on DAVIS and VIP, and saturates on the JHMDB dataset. CDG-MAE trained with 100 epochs outperforms CropMAE even when the latter is trained for longer.

\subsubsection{Training CropMAE with synthetic data}
CropMAE is trained with real images from ImageNet-1K, whereas CDG-MAE is trained with real and diffusion-generated images from ImageNet-1K as pairs of views. In Table~\ref{tab:sup_cropmae_syn}, we present the performance of CropMAE when trained using real and synthetic images as individual data points (Real + Synthetic). Our proposed CDG-MAE outperforms CropMAE (Real + Synthetic) across all three downstream tasks.

\begin{table}[!htbp]
\centering
{\small
\caption{Downstream evaluation of CropMAE, CropMAE (Real+Synthetic), and CDG-MAE.}
\label{tab:sup_cropmae_syn}
\begin{tabular}{c|ccc}
\toprule
   Method & DAVIS & VIP & JHMDB \\
     & $\mathcal{J} \& \mathcal{F}_m$  & mIoU & PCK\@0.1 \\

   \midrule
    CropMAE &  59.7	& 33.8 & 43.9 \\
    CropMAE (Real + Synthetic) & 60.6  & 33.9   & 44.3 \\ 
    CDG-MAE a1  & 61.2   &  37.6  & 46.5  \\ 
    CDG-MAE a3  &  62.6  &  38.1  &  47.8 \\ 
\end{tabular}}
\end{table}

\subsubsection{Scaling}
In this section, we compare scaling the number of patches and model parameters in cross-view MAE. In the main paper section ~\ref{sec:res_scaling} (also presented in Table~\ref{tab:sup_scale_param}), we studied scaling the number of patches from 196 to 784 by decreasing the patch size from 16 to 8. For CDG-MAE, scaling from ViT-S/16 to ViT-S/8 led to large improvements across all three tasks (5.8 on DAVIS, 5.2 on VIP, 9.1 on JHMDB). At ViT-S/8, CDG-MAE outperforms image-based CropMAE and closes the gap to the video-based SiamMAE. We believe that for the downstream tasks of video label propagation, scaling the number of patches has more impact than scaling the number of model parameters. Scaling the number of patches helps models learn features for correspondences under fine-grained changes. We observe that CropMAE does not consistently improve when scaling from ViT-S/16 to ViT-S/8. This is likely because cropped views lack sufficient fine-grained variations to benefit training under a smaller patch size. On the other hand, CDG-MAE and SiamMAE improve since they can learn correspondences under fine-grained changes (pose, motion, viewpoint) between diffusion-generated views (in CDG-MAE) and video frames (in SiamMAE).

We conduct an additional experiment by scaling CDG-MAE from ViT-S/16 to ViT-B/16 and observe that it does not lead to consistent improvements across downstream tasks (Table~\ref{tab:sup_scale_param}). We also report ViT-B/16 performance for SiamMAE and CropMAE. The scaling follows a similar pattern to CDG-MAE on DAVIS and JHMDB. Even for single-image MAE (results of single-image MAE ViT-B/16 and MAE ViT-L/16), performance improves with scale on DAVIS and VIP, but saturates on JHMDB. It is worth noticing that our ViT-S/16 with CDG-MAE performs better than MAE ViT-L/16. Furthermore, previous work addressing cross-view MAE~\citep{gupta2023siamese, eymael2024efficient}, either image- or video-based does not showcase clear and consistent scaling trends in terms of numbers of parameters on video label propagation tasks, which appears to be confirmed in our work. We believe scaling parameters of cross-view MAE methods remains an open research question. It is still not clear whether it comes from data diversity, downstream tasks or, as we also suspect, from additional required tuning (e.g. hyperparameters) to scale such methods.

\begin{table}[!htbp]
\centering
{\small
\caption{Scaling number of patches is more effective than scaling number of parameters. $\dagger$ denotes results from our reproduction. $\ddagger$ denotes results reported from respective papers. }
\label{tab:sup_scale_param}
\begin{tabular}{c|cc|ccc}
\toprule
   Method & Pretraining  &  Arch &  DAVIS & VIP & JHMDB \\
     & Data &  &  $\mathcal{J} \& \mathcal{F}_m$  & mIoU & PCK\@0.1 \\

   \midrule
    CDG-MAE-a3 & ImageNet & ViT-S/16 &  62.6	& 38.1 & 47.8 \\
    CDG-MAE-a3  & ImageNet &  ViT-B/16 &  63.3  & 37.6   & 48.0 \\ 
    CDG-MAE-a3  & ImageNet &  ViT-S/8 &  68.4  & 43.3   & 56.9 \\ 
    
    \midrule
    CropMAE$\dagger$  &  ImageNet & ViT-S/16  & 59.7   &  33.8  & 43.9  \\ 
    CropMAE$\dagger$  &  ImageNet & ViT-S/8 &  58.7  &  32.2  &  55.6 \\ 
    CropMAE$\ddagger$  &  ImageNet & ViT-S/16   & 60.4   &  33.3  & 43.6  \\ 
    CropMAE$\ddagger$  & ImageNet & ViT-B/16  &  60.9  &  32.8  &  44.3 \\ 

    \midrule
    SiamMAE$\ddagger$  &  Kinetics & ViT-S/16  & 62.0   &  37.3  & 47.0  \\ 
    SiamMAE$\ddagger$ &  Kinetics & ViT-B/16 &  62.8  &  38.4  &  47.2 \\ 
    SiamMAE$\ddagger$ &  Kinetics & ViT-S/8  & 71.4   &  45.9  & 61.9  \\ 

    \midrule
    MAE  &  ImageNet & ViT-B/16  & 53.5  &  28.1  & 44.6  \\ 
    MAE  & ImageNet & ViT-L/16 &  56.9  &  29.9  &  44.6 \\

\end{tabular}}
\end{table}

\subsubsection{Video Diffusion Models}
\label{sec:sup_video_diffusion}
We did not experiment with video diffusion models for view generation because they are computationally impractical given our compute budget. SoTA image-to-video models like Wan2.1 ~\citep{wan2025wan} and HunyuanVideo ~\citep{kong2024hunyuanvideo} have substantial computational requirements that make them prohibitively expensive for ImageNet-scale data generation (1.3 M images). Wan2.1’s 14B model requires ~9 minutes to generate a 5 second video on a single A100 GPU. For ImageNet-scale generation, this would require over 1000 days on our 8 * A100 GPU node. Even the more efficient LTX-video ~\citep{hacohen2024ltx} model requires 30 seconds to generate a single video, which would translate to 56 days for ImageNet. However, as video diffusion models become more efficient, we believe they can be easily integrated into our framework. The frames generated by image-to-video diffusion models can be used as a drop-in replacement for the current views generated by image-to-image diffusion models when training CDG-MAE.

\subsubsection{Evaluation on optical flow}

To explore the effectiveness of our method for cross-view tasks beyond label propagation, we evaluate features from CDG-MAE and CropMAE (our primary baseline) on optical flow estimation by training WAFT~\citep{wang2025waft} in Table~\ref{tab:sup_opticalflow}. WAFT utilizes a frozen pretrained encoder as a feature extractor, combined with a trainable recurrent update model implemented via DPT~\citep{ranftl2021vision} to predict optical flow. Following WAFT's standard protocol, we train the models for 50K iterations on the FlyingChairs~\citep{flyingchairs} dataset (Stage 1), followed by 100K iterations on the FlyingThings~\citep{flyingthings} dataset (Stage 2). We then evaluate the models on KITTI~\citep{kitti} in a zero-shot setting. We use the official hyperparameters except for the Stage 2 batch size, which we reduce from 32 to 8 to fit within the GPU memory. The results show that CDG-MAE outperforms CropMAE when evaluated after both Stage 1 and Stage 2, demonstrating the advantages of our pretraining. We are unable to evaluate SiamMAE, as neither the code nor the checkpoints are publicly available.

\begin{table}[!htbp]
\centering
{\small
\caption{Optical flow evaluation on the KITTI dataset. Comparison of CropMAE and CDG-MAE using WAFT training. Performance is evaluated using End-Point Error and the percentage of flow outliers (F1).}
\label{tab:sup_opticalflow}
\begin{tabular}{c|cc}
\toprule
   Method & \multicolumn{2}{c}{KITTI} \\
   &  Endpoint-error ($\downarrow$) & F1 ($\downarrow$)  \\

   \midrule
    CropMAE + WAFT Stage1 &  27.17	& 60.88 \\
    \rowcolor{blue!10}
    CDG-MAE-a3 + WAFT Stage1 &  \textbf{23.77}	& \textbf{52.12} \\
    \midrule
    CropMAE + WAFT Stage2 &  3.81	& 11.75 \\
    \rowcolor{blue!10}
    CDG-MAE-a3 + WAFT Stage2 &  \textbf{3.70}	&  \textbf{11.50} \\
\end{tabular}}
\end{table}

\subsubsection{Consistency metrics with DINOv2 backbone}

In Section~\ref{sec:choice_of_diffusion_model} and Table~\ref{tab:choice_of_diffusion_model}, we evaluated our proposed consistency metrics using an MAE ViT B/16 backbone to determine the optimal diffusion model for bag of views generation. To ensure the robustness of these findings, Table~\ref{tab:choice_of_diffusion_model_wdinov2} reevaluates the same metrics using a DINOv2-S/14~\citep{oquab2023dinov2} backbone. While the absolute metric values differ across backbones, the relative trends remain consistent. The GS, LS, and NPS metrics for views generated by Gen-SIS and Lumos are considerably closer to real video frames than those generated by RCG. Furthermore, although Lumos outperforms Gen SIS on these metrics, and CDG-MAE yields better results when paired with Lumos generated views, we selected Gen-SIS as default choice for bag of views generation in our experiments. This choice allows us to strictly avoid the potential data leakage risks as discussed in Section~\ref{sec:choice_of_diffusion_model}.

\begin{table}[!htbp]
\centering
{
\caption{CDG-MAE (single-anchor setting) performance with different diffusion models (S-LDMs) and corresponding consistency metrics (GS, LS, NPS) evaluated with DINOv2 backbone. Video frame metrics provided as reference. Our proposed consistency metrics strongly effect the performance. Note that the downstream performances are replica of Table~\ref{tab:choice_of_diffusion_model}.}
\label{tab:choice_of_diffusion_model_wdinov2}
\resizebox{\textwidth}{!}{%
\begin{tabular}{c|ccc|ccc}
\toprule
    Diffusion Model & DAVIS  & VIP & JHMDB & Global Sim. ($\uparrow$) & Local Sim. ($\downarrow$)  & Nearest Patch Sim.($\uparrow$ )  \\
    \midrule
    \rowcolor{blue!10}
    Gen-SIS (~\cite{gen-sis}) & 61.2 & 37.6 & 46.5 & 0.779 & 0.555 &  0.753\\
    RCG (~\cite{rcg}) & 57.4 & 34.8 & 43.7 & 0.668 & 0.476 & 0.695 \\
    Lumos (~\cite{ma2025learning}) & 61.9 & 37.7 & 47.3 & 0.824 & 0.608 & 0.792 \\
    \midrule
    Video frames & NA & NA & NA & 0.817 &  0.693 & 0.830 \\
    

\end{tabular}}}
\end{table}

\subsubsection{Quantitative analysis using classification}

In the main text, we analyzed the generated views using our proposed consistency metrics. In this section, we quantify the failure modes of these generated views using a supervised CLIP ViT-B/16 classifier~\citep{cherti2022reproducible}, which achieves a top-1 accuracy of 86.2\% and a top-5 accuracy of 97.8\% on the ImageNet-1K real validation set. Evaluating the classifier on our bag-of-views ($M=4$) across the ImageNet training dataset yields a top-1 accuracy of 73.6\% and a top-5 accuracy of 92.0\%. Notably, 70.9\% of the 1.28M data points have at least 3 out of 4 views correctly classified.

To determine whether classification-based filtering enhances pretraining, we expanded the pool of generated views to $M=8$. For each ImageNet sample, we selectively sampled between 3 and 4 correctly classified views to construct a filtered bag-of-views; this configuration maintains consistency with our original setup of $M=4$ views while utilizing 3 anchor views. Under this constraint, 20.4\% of the original training instances failed to meet the minimum threshold of 3 valid views and were subsequently removed from the pretraining set. We then trained CDG-MAE-a3 with this filtered dataset while keeping the total number of training iterations fixed. As shown in Table~\ref{tab:sup_filter}, this filtering strategy does not yield a performance boost. This outcome suggests that the filtering mechanism undercuts performance by reducing overall data diversity by 20.4\%, or that classification correctness may not align with the optimal filtering criteria for cross-view representation learning.

\begin{table}[!htbp]
\centering
{\small
\caption{Downstream evaluation of our default CDG-MAE-a3 and CDG-MAE-a3 pretrained with classifier-based filtering.}
\label{tab:sup_filter}
\begin{tabular}{c|ccc}
\toprule
   Method & DAVIS & VIP & JHMDB \\
     & $\mathcal{J} \& \mathcal{F}_m$  & mIoU & PCK\@0.1 \\

   \midrule
    \rowcolor{blue!10}
    CDG-MAE-a3  & 62.6  &  38.1  &  47.8 \\  
    w/ filtering  &  62.2  &  37.6  &  47.7 \\ 
\end{tabular}}
\end{table}

\subsubsection{Robustness to bag of views size}

While the main text evaluates configurations with a fixed bag of views size of $M=4$ views per real image, we investigate our method's sensitivity to this parameter in Table~\ref{tab:sub_grid}. Specifically, we evaluate the three anchor configuration by varying $M$, $r_a$, and $r_t$. Our optimal configuration in the main paper involves three anchors, an anchor masking ratio of $r_a=25\%$, and a target masking ratio of $r_t=90\%$, which we compare against settings with $M=6$ and $M=8$.

In Table~\ref{tab:sub_grid}, the settings from the main paper are reported as an average across three random seeds, whereas the exploratory runs are evaluated using a single seed. We observe that performance remains stable across the different choices of $M$, with the results for $M=6$ and $M=8$ remaining close to the default choice of $M=4$. Furthermore, we find that the $r_a = 25\%$ and $r_t = 90\%$ configuration performs best across the different choices of $M$.

\begin{table}[t]
\centering
\caption{Performance of CDG-MAE with three anchors across varying anchor masking ratios ($r_a$), bag of views sizes ($M$), and target masking ratios ($r_t$). Rows reporting standard deviations ($\pm$ std) are averaged over three random seeds.}
\label{tab:sub_grid}

\begin{subtable}[t]{0.45\textwidth}
\caption{Varying $r_a$ and $M$ (with $r_t = 90\%$).}
\centering
\resizebox{0.95\textwidth}{!}{%
\begin{tabular}{cc|ccc}
\toprule
  $r_a$ & $M$ & DAVIS & VIP & JHMDB \\
     & & $\mathcal{J} \& \mathcal{F}_m$  & mIoU & PCK\@0.1 \\

   \midrule
    \multirow{3}{*}{25\%} & 4 & 62.6$\pm$0.1	 & 38.1$\pm$0.1	 & 47.8$\pm$0.2 \\  
     & 6  &  62.4  &  37.9  &  47.8 \\ 
      & 8 &  62.6  &  38.2  &  47.7 \\ 
    \midrule
    \multirow{3}{*}{50\%} & 4 & 62.0$\pm$0.4 & 37.4$\pm$0.3 & 47.5$\pm$0.2 \\  
     & 6  &  61.5  &  37.8  &  47.4 \\ 
      & 8 &  61.7  &  37.4  &  47.4 \\ 
\end{tabular}%
}
\end{subtable}
\hspace{1em}
\begin{subtable}[t]{0.45\textwidth}
\caption{Varying $r_t$ and $M$ (with $r_a = 25\%$).}
\centering
\resizebox{1.0\textwidth}{!}{%
\begin{tabular}{cc|ccc}
\toprule
  $r_t$ & $M$ & DAVIS & VIP & JHMDB \\
     & & $\mathcal{J} \& \mathcal{F}_m$  & mIoU & PCK\@0.1 \\

   \midrule
    \multirow{3}{*}{90\%} & 4 & 62.6$\pm$0.1	 & 38.1$\pm$0.1	 & 47.8$\pm$0.2 \\  
     & 6  &  62.4  &  37.9  &  47.8 \\ 
      & 8 &  62.6  &  38.2  &  47.7 \\ 
    \midrule
    \multirow{3}{*}{98.5\%} & 4 & 60.5  &  35.0  &  45.7 \\  
     & 6  &  60.7  &  34.8  &  45.5 \\ 
      & 8 &  60.6  &  34.8  &  45.5 \\ 
\end{tabular}%
}
\end{subtable}

\end{table}

\end{document}